\documentclass[acmsmall,nonacm]{acmart}
\usepackage{longtable}
\usepackage{afterpage}
\setcopyright{none}
\acmDOI{}
\acmISBN{}
\graphicspath{{figures/}}

\title[EMS-Nav]{EMS-Nav: Embodied Multi-Scale Navigation}

\author{Zenghui Yang}
\affiliation{\institution{Tsinghua University}\city{Beijing}\country{China}}
\email{zenghuiyang36@gmail.com}

\author{Xinan Wang}
\affiliation{\institution{Tsinghua University}\city{Beijing}\country{China}}
\email{wxa25@mails.tsinghua.edu.cn}

\author{Danyang Li}
\affiliation{\department{School of Software}\institution{Tsinghua University}\city{Beijing}\country{China}}
\email{thuldy@mail.tsinghua.edu.cn}

\author{Zhe Xu}
\affiliation{\department{School of Software}\institution{Tsinghua University}\city{Beijing}\country{China}}
\email{z-xu26@mails.tsinghua.edu.cn}

\author{Zheng Yang}
\authornote{Corresponding author.}
\affiliation{\institution{Tsinghua University}\city{Beijing}\country{China}}
\email{yangzheng@tsinghua.edu.cn}

\author{Jingao Xu}
\affiliation{\department{Electrical and Computer Engineering}\institution{The University of Hong Kong}\city{Hong Kong SAR}\country{China}}
\email{jingaoxu@hku.hk}

\author{Qiang Ma}
\affiliation{\institution{Qiyuan Lab}\city{Beijing}\country{China}}
\email{maqiang@qiyuanlab.com}

\begin{document}\hypersetup{hidelinks}
\begin{abstract}
Applications such as industrial inspection require robots to autonomously reach different work areas and specific pieces of equipment. These destinations span spatial scales from regions to objects, so robots must relate each target to its surroundings to interpret the task and determine where to go. Existing methods support multi-scale navigation through rich environment representations, but at substantial computational and storage cost. These costs limit deployment where wireless coverage is incomplete and robots must rely on onboard computation to keep navigating. We introduce EMS-Nav, a biologically inspired multi-scale navigation framework for online, on-device operation. Its core innovation is to organize perception, spatial understanding, and navigation around regions, making local spaces the basic units of both environmental understanding and computation. EMS-Nav groups objects and their relationships into functionally meaningful spaces and builds a hierarchy from local areas to the broader environment. This organization supports efficient map construction alongside the interpretation and execution of language goals at different scales. In multi-scale navigation experiments, EMS-Nav achieves the highest success rates across all four target categories: objects, rooms, regions, and specific instances. It outperforms representative state-of-the-art open-vocabulary mapping and navigation methods, improving overall first-attempt success by 9.39 percentage points over the strongest baseline while reducing mapping overhead and language-model input size. Local deployment in real indoor environments further demonstrates its applicability to on-device embodied navigation. By establishing regions as the organizing basis for perception, spatial understanding, and language-guided action, EMS-Nav provides a system that combines multi-scale understanding with computational efficiency to support diverse language goals on resource-constrained robots.

\end{abstract}
\maketitle

\afterpage{
\begin{figure}[!t]
  \centering
  \includegraphics[width=\textwidth,pagebox=cropbox]{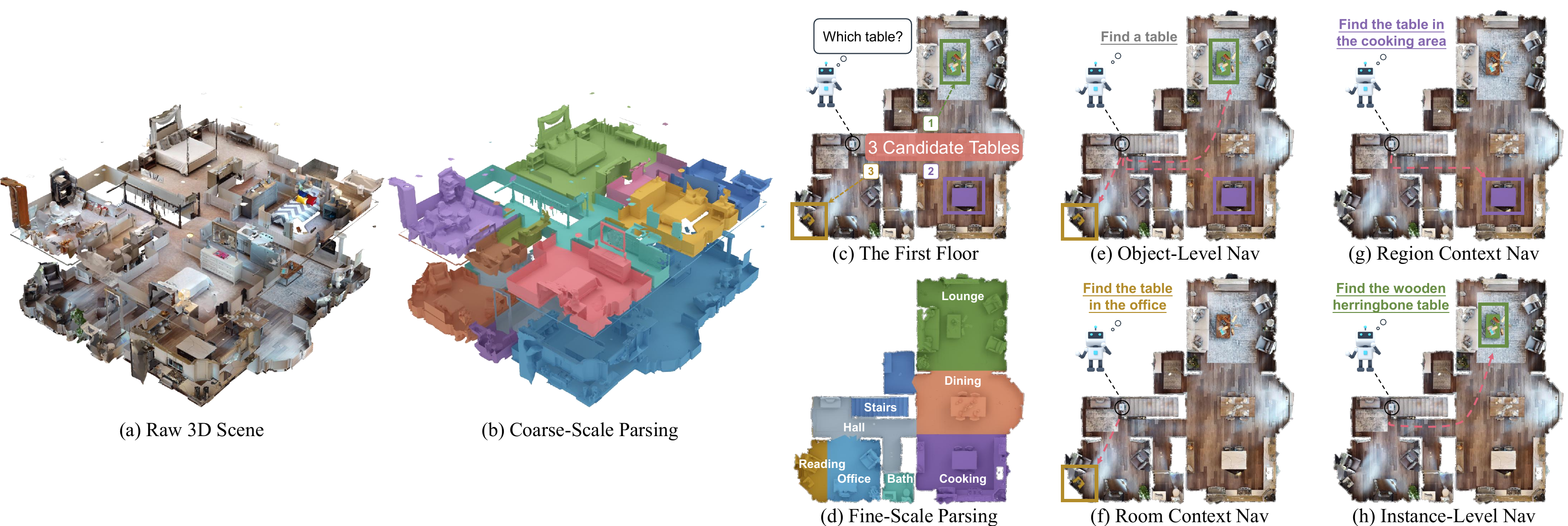}
  \caption{Region-centered navigation across spatial scales. (a,b) The scene and its coarse spatial partition. (c,d) Three candidate tables and their local functional regions. (e--h) Language-guided navigation tasks with object, room, region, and instance constraints. Region membership and hierarchical links connect objects to their surrounding spaces, enabling navigation to destinations specified at different scales within the same environment.}
  \Description{The left panels show the original 3D scene and its coarse partition. The middle panels show three candidate tables on the first floor and fine-scale regions. The right panels illustrate targets and navigation for four types of language instructions.}
  \label{fig:teaser}
\end{figure}
}

\section{Introduction}
\label{sec:intro}

Navigation is a fundamental capability of embodied robots, connecting perception of the physical world to action within it. Mobile robots support warehouse logistics~\cite{amazon2022navigation}, hospital deliveries~\cite{toyota2026potaro}, industrial inspections~\cite{anybotics2025deployment}, and so on. These tasks require robots to identify destinations and navigate to them. When tasks are specified in natural language, destinations may be described at different spatial scales.

A user may ask a robot to reach a room, a functional region, or an object, or combine these scales in one instruction. For example, ``Find the laptop on the mint-green computer desk in the office'' identifies the target through its room, desk attributes, and spatial relationship to the desk. Recognizing the laptop alone is insufficient: multi-scale semantic navigation requires connecting object details with their surrounding spaces~\cite{hovsg,miao2026langmap}.

Existing studies support such multi-scale semantic navigation through hierarchical representations and spatial context. HOV-SG~\cite{hovsg} organizes floors, rooms, and objects to localize language goals, while SG-Nav~\cite{yin2024sgnav} uses relationships among objects, object groups, and rooms for object-goal search. LangMap~\cite{miao2026langmap} evaluates goals at the scene, room, region, and instance levels.

However, constructing rich environment representations can incur substantial computation and storage costs~\cite{conceptgraphs,hovsg}. Where wireless coverage is incomplete, as in industrial inspection~\cite{anybotics2023connectivity}, robots must continue navigating with limited onboard compute and memory, constraining the deployment of these methods.

To reduce these costs, other work explores more efficient map construction and use. Clio~\cite{clio} builds compact online scene graphs by retaining task-relevant objects and regions. OneMap~\cite{onemap} reuses semantic maps for successive object searches on embedded hardware, while FindAnything~\cite{findanything} improves object semantic map construction and storage efficiency.

Despite these advances in semantic representation and efficiency, the object-level scene graph methods above largely follow an object-first paradigm: they first reconstruct individual objects, then use those representations to infer spatial relationships and higher-level structure. Separating object reconstruction from spatial organization creates two limitations for on-device multi-scale navigation: the cost of processing a growing body of object information, and the difficulty of recovering functional spaces from object representations.

\noindent $\bullet$\enspace Building and interpreting an environment around objects places increasing computational demands on the system as the scene grows~\cite{conceptgraphs,hovsg}. As exploration extends, the map accumulates more objects and relationships. If a local change triggers reprocessing across a wider area, map maintenance and spatial interpretation repeatedly revisit unrelated object details. This mismatch consumes scarce on-device resources and limits efficient online mapping and spatial understanding as coverage expands.

\noindent $\bullet$\enspace Object reconstruction may leave navigation-relevant spatial structure implicit. A laboratory's preparation, storage, and measurement areas can occupy the same room and contain similar object categories. Their differences depend on local layouts, boundaries, and arrangements of objects. Room boundaries alone cannot distinguish these areas, and object labels alone cannot determine their extent~\cite{hydra,yin2024sgnav}. The same function can also occur in different layouts, limiting the usefulness of fixed spatial partitions. Thus, even when a target object is already mapped, a robot may be unable to determine whether it lies in the functional space named by the instruction. Resolving this ambiguity requires a map that connects objects to the functional spaces they collectively form and represents hierarchical relationships among those spaces.

Addressing these limitations poses three connected challenges for on-device multi-scale navigation. The first is to make sense of local space from partial observations: objects may be recognizable, but which objects form a coherent space, and how they should be organized, are not given. The second is to derive spatial hierarchies that align with everyday understanding and language. Functional relationships need not follow walls or fixed architectural levels, limiting predefined partitions. The third is to retrieve goal-relevant spatial knowledge under limited resources without processing every scale in each decision.

When people seek a destination, they connect visible objects to their surroundings and draw on spatial knowledge at the scales relevant to the description. This motivates us to look to biological spatial cognition for design inspiration. Studies of local layout perception~\cite{epstein1998local}, hippocampal cognitive maps~\cite{iaria2007maps}, and prefrontal goal and contextual control~\cite{koechlin2003control} suggest a functional division among local perception, spatial integration, and goal-directed decisions. We draw from this division a design principle: use local spaces to connect object perception with understanding of the broader environment, and access different levels of spatial knowledge according to the navigation goal. This leads us to place regions at the center of perception, understanding, and navigation.

We introduce EMS-Nav, a multi-scale language-guided navigation framework designed for online, on-device operation. Its core innovation is a region-centered organization that extends spatial structure beyond an interpretation added after object reconstruction, making it foundational to perception, understanding, and navigation (Figure~\ref{fig:teaser}).

EMS-Nav addresses the three challenges within a shared spatial organization. Perceptual structuring inspired by the scene-selective cortex (SSC) forms spatial units from local layouts and object relationships, establishing how the local environment is represented. Hippocampus-inspired spatial parsing constructs a multi-scale hierarchy from functional relationships, supporting spatial understanding beyond fixed room partitions. Goal-directed navigation inspired by the prefrontal cortex (PFC) selects knowledge at the scales relevant to an instruction and uses it to determine the destination. All three modules organize computation around regions: local changes are updated primarily within the relevant spaces, while broader understanding builds on region summaries. Resource constraints thus inform perception, parsing, and navigation throughout the framework.

Our contributions are threefold:
\begin{itemize}
\item We introduce EMS-Nav, a multi-scale language-guided navigation framework for online, on-device operation. It connects environmental mapping, spatial understanding, and language-guided goal decisions, enabling robots to use a shared map for navigation instructions at different scales under limited computational resources.
\item We propose a spatial organization that uses regions to connect perception with semantic parsing, making local spaces the basic units of both environmental understanding and computation. By extending object understanding to functional regions and their hierarchical relationships, this organization allows multi-scale spatial understanding and efficient mapping to support one another.
\item Large-scale evaluation and real-world deployment demonstrate improvements in both multi-scale navigation and computational efficiency. Across 18,628 instructions covering object, room, region, and specific-instance targets, EMS-Nav achieves the highest navigation success rates in all four categories, outperforming representative state-of-the-art open-vocabulary mapping and navigation methods while reducing mapping overhead and language-model input size. Real-world deployment on an RTX 4060 Ti reaches a peak on-device processing throughput of 5.19 frames per second (FPS), further demonstrating the framework's practical applicability. We will release the source code for EMS-Nav.
\end{itemize}

\section{Related Work}
\label{sec:related-work}

\subsection{Language Goals and Multi-Scale Semantic Navigation}

Language-guided navigation requires robots to ground human descriptions in actionable goals in their environment. R2R pairs natural-language route instructions with panoramic trajectories through real indoor environments\cite{anderson2018r2r}, while RxR extends this setting with multilingual instructions and spatiotemporal alignments\cite{ku2020rxr}. VLN-CE extends instruction following to low-level actions in continuous environments~\cite{krantz2020vlnce}. These tasks primarily assess the ability to follow verbally described routes. Object-goal navigation instead focuses on finding a specified object; GOAT-Bench combines category, language, and image goals in sequential tasks\cite{khanna2024goatbench}. LangMap's HieraNav further evaluates the understanding of spatial constraints through scene-, room-, region-, and instance-level goals\cite{miao2026langmap}. Together, these tasks extend navigation requirements from object names to richer spatial contexts. A map must therefore support queries about the space containing a target and its surrounding context, as well as the target itself.

Foundation-model navigation systems also differ in how they connect language to action. LM-Nav composes pretrained language, visual-language, and navigation models to ground instructions in a topological map~\cite{shah2023lmnav}. NavGPT reasons over textual observations and navigation history~\cite{zhou2024navgpt}, while NavGPT-2 combines visually grounded language reasoning with a learned policy~\cite{zhou2025navgpt2}. NaVid instead predicts actions from instruction-conditioned RGB video history~\cite{zhang2024navid}. These approaches demonstrate that language-guided behavior can draw on different intermediate representations and observation histories. Our focus is the construction of spatial evidence itself: maintaining an environment representation in which object details remain connected to functional regions and larger spatial groupings, so that different navigation goals can query the relevant context.

\subsection{Multi-Scale Spatial Representations and Scene Graphs}

Spatial representations connect local detail with the broader organization of an environment. Earlier semantic and instance mapping fused predictions across views~\cite{mccormac2017semanticfusion,runz2018maskfusion,mccormac2018fusionpp,narita2019panopticfusion}. Early 3D scene graphs unified objects, spaces, and camera information\cite{armeni3dsg}; Kimera and Hydra subsequently organized geometry, objects, places, and rooms into incrementally constructed hierarchies\cite{kimera,hydra}. Learning-based 3D scene graph prediction learns object relations from reconstructed data\cite{sg3d2020}, and SceneGraphFusion incrementally predicts objects and relations from RGB-D sequences\cite{scenegraphfusion}. ConceptGraphs builds open-vocabulary object graphs through geometric and semantic association across views, using language models to generate object relations\cite{conceptgraphs}. OpenMap uses structural--semantic consensus to aggregate instances and grounds instructions using spatial context~\cite{li2025openmap}. HOV-SG organizes floors, rooms, and objects into a hierarchy for progressively localizing complex language goals\cite{hovsg}. We likewise organize maps hierarchically, using ``multi-scale'' to describe access to local nodes, functional regions, and larger spatial groupings. Higher-level regions emerge through spatial and functional connections, without fixing hierarchy depth to architectural categories.

Region semantics enrich object-level representations with local function and spatial context. SG-Nav introduces groups of related objects between objects and rooms, such as a dining table and its surrounding chairs, using object relations and a predefined dictionary of category associations\cite{yin2024sgnav}. GLMap jointly maintains instance and region semantic units, providing both text and renderable visual information\cite{glmap}. These designs show that the organization of objects itself provides evidence for navigation. Yet architectural topology does not always determine region boundaries: Hydra explicitly discusses the difficulty of separating semantically distinct spaces in open layouts using topology-based room detection alone\cite{hydra}. EMS-Nav uses regions for both map formation and spatial interpretation. Coarse regions constrain candidates for subsequent superpoint association and node merging, while periodic parsing refines these regions and builds larger-scale semantic organization.

Open-vocabulary 3D representations provide a more flexible semantic basis for language queries. ConceptFusion fuses multimodal features into 3D maps\cite{conceptfusion}, OpenScene supports open-category queries through correspondences between 2D and 3D features\cite{openscene}, and VLMaps associates visual-language features with map locations for navigation\cite{vlmaps}. OpenMask3D aggregates instance-level features across views~\cite{takmaz2023openmask3d}, while Open3DIS uses 2D masks to guide open-vocabulary 3D instance segmentation~\cite{nguyen2024open3dis}. Another line of work embeds language features in renderable scenes: LERF uses radiance fields\cite{lerf}, whereas LangSplat uses 3D Gaussians with language features\cite{langsplat}. These representations make scene details accessible through language, but open-vocabulary querying does not itself establish functional regions or their hierarchical organization. EMS-Nav retains object-level evidence while also organizing the spaces that those objects collectively form, connecting access to individual objects with access to their broader spatial context.

\subsection{Efficient Mapping and Parsing under Resource Constraints}

Continuous mapping in embodied settings must manage the processing burden of accumulating observations. Existing approaches make different tradeoffs among representational granularity, task information, and resource use. Open-Fusion integrates open-vocabulary semantics into real-time 3D mapping\cite{openfusion}. Clio uses an information bottleneck to aggregate objects and spatial primitives according to predefined language tasks, constructing compact hierarchical graphs with onboard computation\cite{clio}. OneMap reuses uncertainty-aware feature maps across sequential object searches and runs on embedded hardware\cite{onemap}. FindAnything integrates object- or part-level features into volumetric submaps, controlling costs through association and aggregation within the active submap\cite{findanything}. DualMap combines abstract and concrete maps and handles changes through lightweight object-state checks\cite{dualmap}. Compact segmentation models offer a complementary component-level approach to efficiency, including MobileSAM, EfficientSAM, and EfficientViT-SAM~\cite{zhang2023mobilesam,xiong2024efficientsam,zhang2024efficientvitsam}. EMS-Nav organizes continuous updates around local regions and constructs functional spaces through periodic parsing, controlling update scope while retaining multi-scale semantic information.

Reducing the information read at query time offers another way to lower navigation reasoning costs. SayPlan searches a collapsed scene graph for task-relevant subgraphs and combines them with a classical planner\cite{sayplan}. KeySG selects keyframes that cover the environment, summarizes rooms and floors hierarchically, and retrieves relevant context through hierarchical multimodal retrieval\cite{keysg}. Memory Over Maps stores posed RGB-D keyframes, retrieves views at query time, and estimates target 3D locations on demand, reducing upfront reconstruction\cite{memoryovermaps}. These approaches adjust either the scope of information accessed or the information built in advance. EMS-Nav uses regions to organize both representation construction and subsequent access: online updates select local candidates, while periodic parsing recomputes changed regions and their affected ancestors and reuses descriptions of unchanged regions.

\subsection{Open-Vocabulary and Language-Guided Navigation}

Object search in unknown environments also requires deciding where to gather information next. SemExp combines semantic mapping with an exploration policy\cite{chaplot2020semexp}, while PONI uses potentials over objects and unexplored areas to guide search\cite{ramakrishnan2022poni}. For open-vocabulary goals, ZSON and CoW use pretrained visual-language representations for transfer or goal localization\cite{majumdar2022zson,gadre2023cow}. L3MVN, LFG, and ESC incorporate language knowledge or commonsense constraints into frontier selection\cite{yu2023l3mvn,shah2023lfg,zhou2023esc}; VLFM constructs a value map from image--goal-text similarity\cite{yokoyama2024vlfm}. VoroNav uses paths and semantic descriptions from a reduced Voronoi graph to select exploratory waypoints~\cite{wu2024voronav}. These methods primarily address search and exploration decisions in unknown environments. Our work focuses on organizing accumulated observations across spatial scales and evaluates the localization of language goals and navigation to them within the resulting map. This setting isolates how the organization of observed space supports goal interpretation.

Scene graphs provide structured evidence beyond individual objects for navigation reasoning. SG-Nav reasons hierarchically over objects, object groups, and rooms, and uses re-perception to address false-positive target detections\cite{yin2024sgnav}. Wu et al. aggregate objects, local areas, and regions online, combining semantic priors with exploration evidence and using graph-based simulation for long-horizon planning\cite{wu2026hsg}. Unlike these approaches to exploration in unknown environments, we first construct a representation along a given trajectory and then evaluate multi-scale goals using LangMap instructions\cite{miao2026langmap}. EMS-Nav preserves region membership and links between hierarchy levels, allowing target selection to combine broader spatial context with local object detail. All methods in our experiments receive the same observations and attempt budgets, enabling a comparison of the role of spatial representations in this task.

\section{Method}
\label{sec:method}
\subsection{Framework Overview and Representation}
\label{sec:method-overview}

EMS-Nav maintains two related spatial organizations. Online coarse regions group map nodes to limit association and merging candidates during map construction. Periodic parsing refines these groups into functional regions and composes them into a semantic hierarchy for language-guided target selection. The hierarchy is derived from the online map and coarse partition; it does not replace the coarse partition used by subsequent mapping updates.

EMS-Nav organizes the environment around regions, allowing local spaces to guide object map construction while connecting object details to an understanding of the wider environment.  Given posed RGB-D observations, the perception frontend extracts superpoints with geometric and visual features and projects them into world coordinates.  Each node in the online map $G_t$ stores either a historical superpoint or an aggregate of superpoints; a map node therefore need not represent a fully reconstructed object instance. The system also maintains a node relation graph $R_t$, a coarse region partition $C_t$, and the observed structural geometry $B_t$. Coarse regions organize online instance association. Periodic spatial parsing produces a semantic hierarchy $H$ containing region descriptions, links to subregions, and membership links to underlying nodes. The navigation module uses this hierarchy to relate the spatial context in an instruction to a specific target, grounding language descriptions at different scales in destinations on the map.

\begin{figure}[tbp]
  \centering
  \includegraphics[width=\linewidth,pagebox=cropbox]{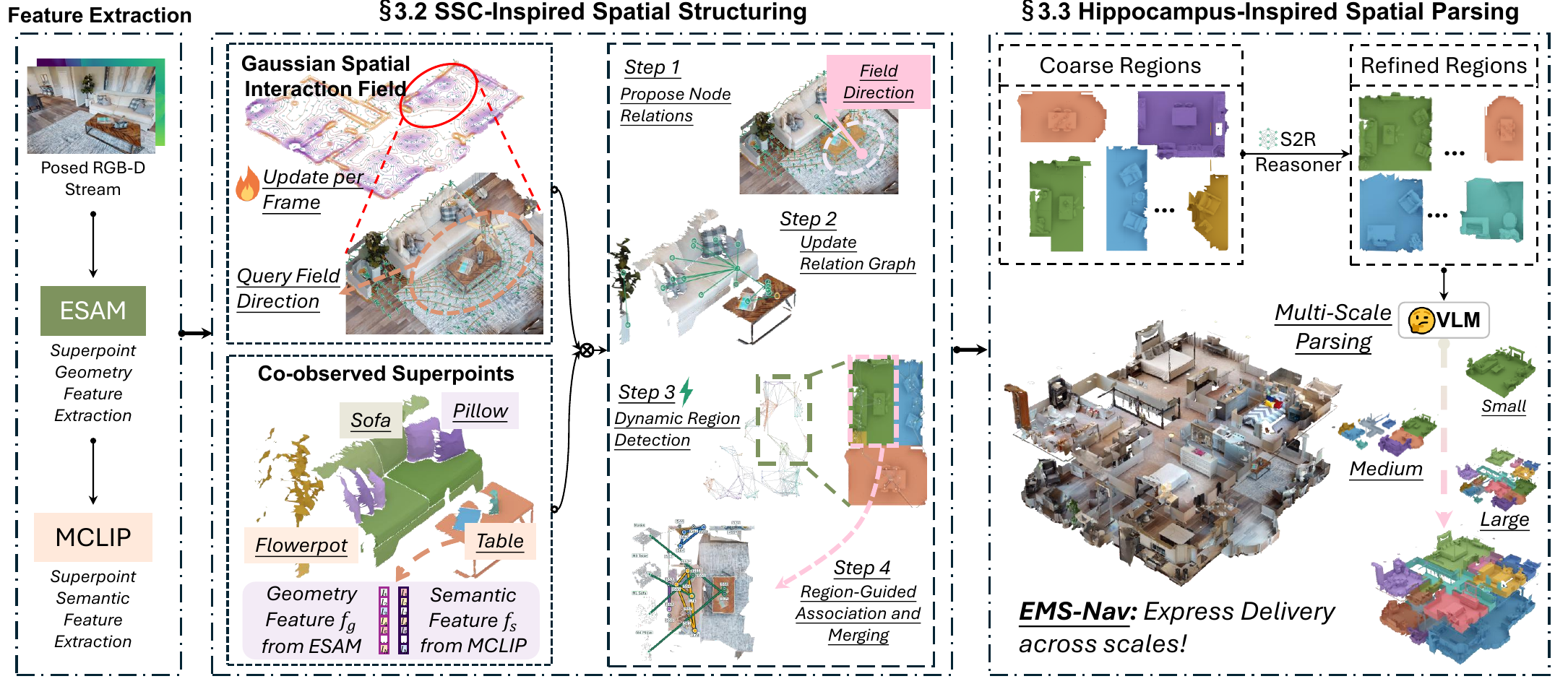}
  \caption{EMS-Nav uses regions to connect online mapping with multi-scale spatial parsing. Left: posed RGB-D observations provide superpoint geometry and visual features. Center: GSIF-guided relations form coarse regions that constrain instance association and merging. Right: periodic S2R refinement and vision-language parsing build functional regions and their semantic hierarchy for navigation. Spatial parsing leaves the online coarse partition unchanged. MCLIP denotes MobileCLIP-S2.}
  \Description{Posed RGB-D observations pass through superpoint feature extraction, Gaussian fields, a relation graph, and online regions to local instance merging. A periodic branch refines the regions and performs multi-scale spatial parsing.}
  \label{fig:method-overview}
\end{figure}

\begin{figure}[t]
  \centering
  \includegraphics[width=\linewidth,pagebox=cropbox]{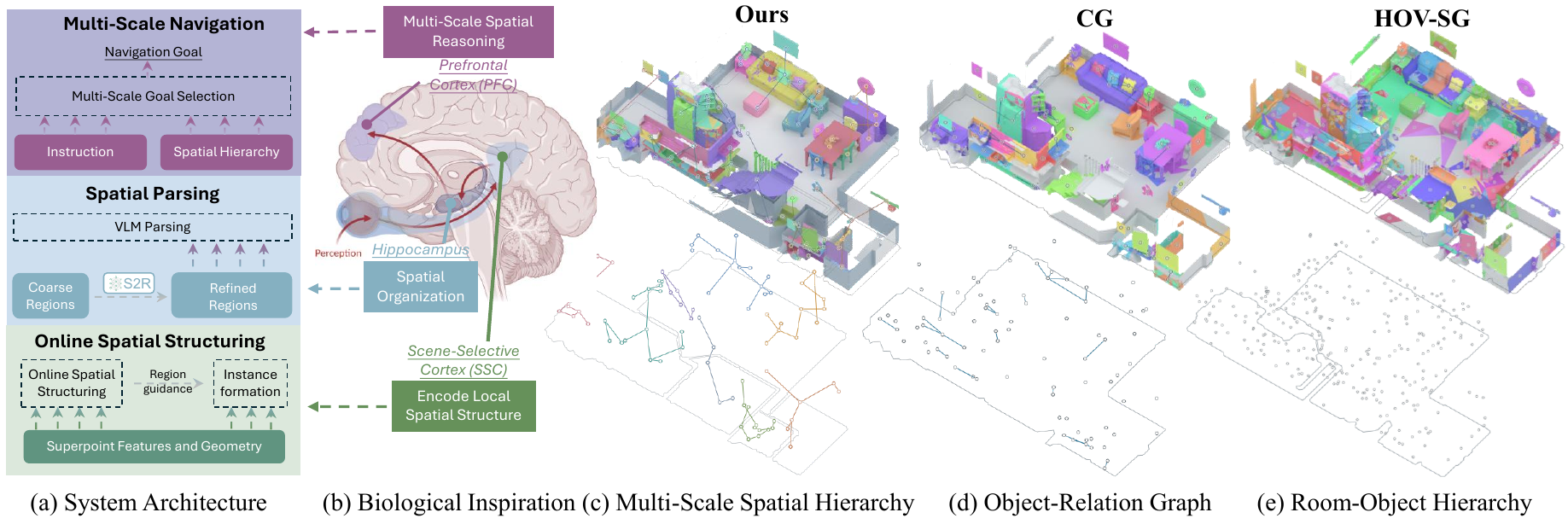}
  \caption{System architecture and spatial organization. (a,b) Perception, spatial parsing, and navigation draw inspiration from local layout perception, spatial integration, and goal-directed control. (c) Nodes and edges within a functional region share a color; translucent outlines indicate structures one level above. (d,e) CG links objects through relations, while HOV-SG groups them within rooms. EMS-Nav connects functional regions to larger spaces for multi-scale goal interpretation.}
  \Description{The left side shows biological inspiration and the division of responsibilities among perception, local regions, and semantic parsing. The right side compares region-guided mapping with other forms of spatial organization.}
  \label{fig:method-architecture}
\end{figure}

Figure~\ref{fig:method-architecture} illustrates how regions connect local perception, spatial integration, and goal selection. The design draws on scene-selective cortex (SSC) for organizing evidence about local layouts~\cite{epstein1998local}, and on the hippocampus for integrating local spaces~\cite{iaria2007maps}. Online spatial structuring maintains coarse regions from geometric, visual, and co-occurrence evidence and uses them to guide instance updates. Periodic spatial parsing reads the current map and partition, refines the regions, interprets their functions, and constructs a semantic hierarchy. The PFC-inspired multi-scale goal-directed navigation module in Section~\ref{sec:navigation} then selects region context, object attributes, and spatial relations according to the instruction. Regions and membership links connect the three modules, so that local organization in the observations becomes the basis for understanding functional spaces and selecting navigation targets.

Figure~\ref{fig:method-architecture}(c--e) illustrates the spatial context made explicit by the representation. In EMS-Nav, functional regions distinguish local uses within a larger space while retaining links to their member nodes and higher-level regions.  These links let navigation access local composition and broader context together.

Online updates and periodic parsing operate at different granularities, as shown in Figure~\ref{fig:method-overview}. For each frame, the system first selects candidate map nodes using the previously available coarse regions, then fuses superpoints, inserts new nodes, and updates relations. Changes to the relation graph drive updates to the coarse partition, supplying candidates for the next observation. At a configurable frame interval, spatial parsing is triggered in parallel with online mapping. It reads the current map and partition to perform within-region subdivision, learned merging, and multi-scale semantic parsing. The semantic hierarchy updates the interpretations of changed regions and affected parent regions, while retaining the results for other regions. The online coarse partition continues to evolve independently. Regions thus constrain instance association and allow spatial parsing to reuse unchanged results, concentrating computation on relevant local spaces and affected higher-level interpretations. Continuous mapping and multi-scale understanding share local spatial information while avoiding repeated processing of unrelated areas.

\subsection{SSC-Inspired Spatial Structuring}
\label{sec:ssc}
\subsubsection{Frame-Level Observations and Local Layout Evidence}

Organizing the environment into regions begins with relating objects to their surroundings from partial observations. Superpoints in the same frame carry both their individual appearance and a shared visibility context. Their shapes, relative positions, and nearby structures such as walls provide further clues about the local layout. The left side of Figure~\ref{fig:observation-layout} juxtaposes a top-down view with observations from different viewpoints, illustrating how objects in a local space can be repeatedly observed together as the viewpoint changes. A map that retains only aggregated object features and positions leaves this co-visibility evidence from map construction difficult to access during subsequent region parsing. EMS-Nav therefore proposes local relations during perception and accumulates spatial connections from co-observations in the map, providing evidence for region formation and subsequent functional interpretation.

Co-observation must be combined with layout and visual semantics to provide a stronger basis for region organization. For example, objects in a bedroom and those in a small adjoining bathroom may rarely appear together in a local observation because of walls, doorways, and viewpoint constraints. Conversely, objects in different functional regions may be visible together through a doorway. We therefore construct relations from co-observation, spatial boundaries, and visual features, rather than assigning all superpoints in a frame to one region. The layout cues in Figure~\ref{fig:observation-layout} identify spatially supported connections within observations. They complement the semantic similarity supplied by CLIP-style vision-language features~\cite{clip2021}, allowing node relations to reflect both semantic affinity and spatial layout as a basis for region partitioning.

\subsubsection{Relation Construction Guided by a Gaussian Spatial Interaction Field}

\begin{figure}[t]
  \centering
  \includegraphics[width=\linewidth,pagebox=cropbox]{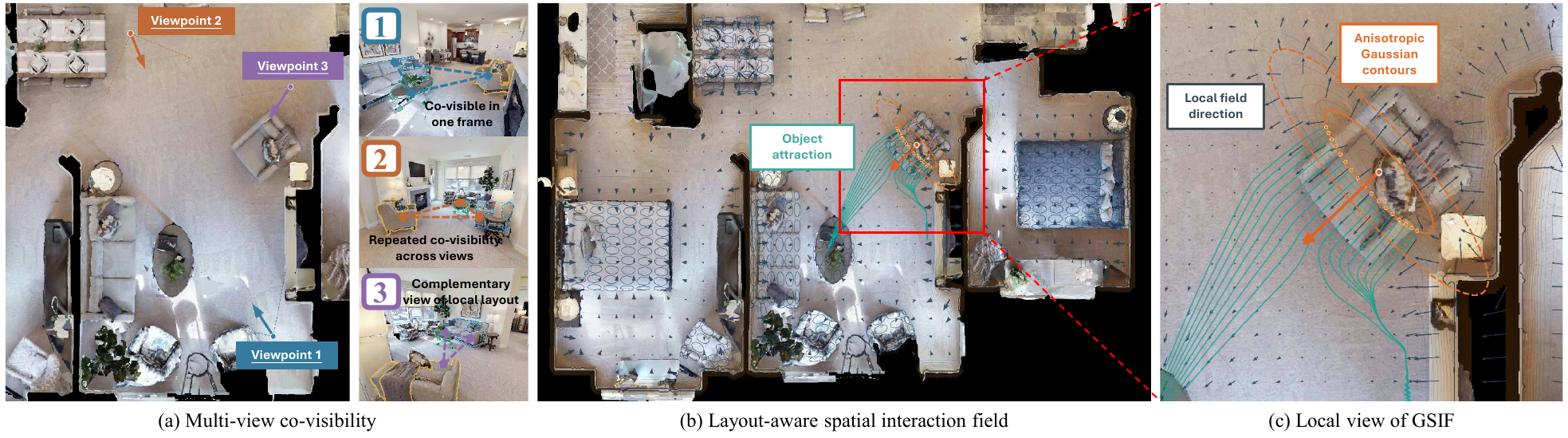}
  \caption{Observation evidence for organizing local spaces. (a) A top-down view and three viewpoints illustrate repeated co-visibility among local objects. (b) GSIF combines object geometry and architectural boundaries to guide relation proposals. (c) Enlarged Gaussian contours and field directions show how shapes and surrounding layout influence candidate connections. These cues complement visual semantics for region formation; co-visibility alone does not determine membership.}
  \Description{The left side shows a local space and three observation viewpoints. The center shows the top-down scene layout and field directions. Enlargements on the right show Gaussian shapes and directional cues around objects.}
  \label{fig:observation-layout}
\end{figure}

We introduce a Gaussian Spatial Interaction Field (GSIF) to guide local relation proposals using object shapes and architectural boundaries. Anisotropic Gaussian kernels approximate the observed horizontal shapes of superpoints, preserving their principal orientations and extents. When querying a connection direction, local repulsion from the source object acts together with attraction from other objects and repulsion from fixed architectural structures such as walls, directing connections outward from the source. The geometric representations are continuously updated as observations accumulate. At locations where relations are needed, the system superposes local contributions and reads the direction of field variation. Thus, even for two nodes that are close in Euclidean distance, an observed wall between them or the direction in which an object extends can affect the proposed relation. Figure~\ref{fig:observation-layout}(b,c) relates object contours and architectural boundaries to local field directions, illustrating how GSIF brings layout evidence into relation proposals.

GSIF guides the system to propose a small set of spatially supported node relations from the current observation and accumulate these proposals across frames in the relation graph $R_t$. Among nearby co-observed nodes, the system preferentially nominates those lying along the locally guided field direction, then selects sparse connections according to a GSIF cost. This cost measures how well a connection agrees with the field guidance. Edge weights combine visual similarity, spatial distance, and support across frames; Appendix~\ref{app:gsif-relations} gives the definitions. This process turns local layout evidence from co-observations into persistent node relations. The same relations support online region partitioning and provide neighborhood context to S2R, allowing mapping and subsequent region understanding to share local spatial connections.

Functional grouping must account for how objects share a space, as well as how they differ in appearance and category. A workbench and a beaker, for example, differ in appearance and category but may jointly form a preparation area. Conversely, beakers in different preparation areas should not be grouped into one space solely because they are semantically similar. The co-observation and layout cues in Figure~\ref{fig:observation-layout} provide evidence beyond individual object attributes: co-observation records which nodes have appeared in the same view, GSIF uses object shapes and architectural boundaries to select connections, and visual features contribute semantic relations between observations. Together, these cues enter the relation graph, allowing subsequent grouping to interpret node relations in their local spatial context. Region organization thus draws on both layout and semantics from the perception stage onward, providing a basis for understanding different objects as parts of one functional space.

\subsubsection{Maintaining Coarse Regions Online}

To keep regions involved throughout map construction, the system updates local spatial groups as observations arrive using Dynamic Frontier (DF) Louvain~\cite{dflouvain2024}, which reuses the existing partition. Node insertion, fusion, and merging, together with changes in observed relations, generate updates to the relation graph. Starting from the existing community assignment, the system adjusts the parts affected by these changes to obtain the coarse partition $C_t$. This process corresponds to dynamic region detection in Figure~\ref{fig:method-overview}. Coarse regions retain their member nodes and internal relations, organizing accumulated spatial connections into candidate sets for subsequent instance association. Reusing the existing partition reduces repeated processing of unaffected parts of the map.

\subsubsection{Region-Guided Superpoint Association and Node Merging}

Coarse regions bring local spatial organization into the next observation update by limiting candidate map nodes. For each current superpoint, the system selects relevant regions within $1\,\mathrm{m}$ of its bounding box (bbox) and collects their member nodes for comparison. A superpoint may select multiple regions, allowing its candidate set to cross existing partition boundaries. Geometric and feature consistency still determine the final match. Regions therefore localize instance comparisons to spaces relevant to the current observation while allowing matches across region boundaries. The connection from dynamic region detection to local instance merging in Figure~\ref{fig:method-overview} illustrates how spatial organization directly guides object map construction.

Instance updates and region maintenance interact as observations accumulate, allowing local spatial organization to guide the map as it takes shape. Initially, with no existing map nodes, all superpoints are inserted into the map. Relations and regions are then established for use in the next observation. In later frames, successfully matched superpoints are fused into historical nodes. Superpoints with no available region or no consistent match become new nodes that can be associated in future observations. On a separate schedule, the system also collects recent observations and merges nodes that may represent the same object within candidate sets constrained by coarse regions, using geometric overlap, feature consistency, and cross-frame association support. The updated nodes and relations in turn maintain the coarse regions. Across frames, observations establish local organization, which then guides instance updates. Regions can thus participate in mapping before objects are fully reconstructed, while accumulating nodes and layout relations for subsequent spatial parsing.

\subsection{Hippocampus-Inspired Spatial Parsing}
\label{sec:spatial-parsing}
\subsubsection{Periodic Subdivision and Learned Region Merging}

Online coarse regions organize candidates for instance association; understanding functional spaces additionally requires determining which nodes jointly form a meaningful region. The first DF Louvain partition serves online candidate selection. At a configurable fixed frame interval, periodic spatial parsing applies Louvain~\cite{louvain2008} once to the induced subgraph of each changed coarse region, producing smaller subregions. This subdivision opens up the interior of a coarse region to provide units that learned reasoning can recombine. A Superpoint-to-Region Reasoner (S2R) scores whether two candidate subregions should merge, using the map-node tokens belonging to each. Grouping subregions along accepted pairs produces functional regions; S2R is trained to approximate the manually checked partitions. Local groupings that support online mapping thereby become the basis for understanding functional spaces.

S2R learns how objects jointly form local functional spaces from manually checked functional region annotations. We use a large model to obtain coarse region annotations for instances in 1513 ScanNet scenes~\cite{scannet2017}, followed by manual checking to obtain spatial groupings judged reasonable by humans. The model learns to organize existing map nodes into regions that approximate these annotations. Each node may contain either a single historical superpoint or an aggregate of historical superpoints. Ideally, the union of the point clouds of all nodes in a predicted region matches the union of the point clouds of all instances in a manually annotated functional region. Point-cloud correspondences transfer region membership supervision from these annotations to map nodes. This connects instance-based functional groupings to the nodes that are still being accumulated and fused during online mapping.

S2R processes local candidate batches to determine whether the subdivided regions should be merged. Each participating map node corresponds to a token, and these nodes form the input units for the candidate subregions. An accepted combination retains all member nodes. Unmerged subregions remain in the map, including singleton groups for which there is insufficient evidence to join a neighboring region. Local batches restrict merging decisions to the candidate set, avoiding the need to pass every node in the scene through the model at once. The transition from coarse to fine-grained regions on the right side of Figure~\ref{fig:method-overview} illustrates this reorganization around functional composition. The resulting regions support functional descriptions and spatial integration at larger scales.

Each node token combines the node's own observations with context from its direct GSIF-guided neighbors. The node's observation features, three-dimensional center, and scale describe its observed content, position, and extent. Along the sparse connections selected from same-frame co-observations and accumulated in the map, the model aggregates local context using neighbor features, relative positions, and GSIF connection costs. The endpoints of a historical connection need not remain jointly visible in the current frame. GSIF thus supports online region partitioning and brings local composition from previous observations into the S2R representation through neighborhood context.

A region is represented by the set of tokens of its member nodes, preserving their observed content, geometry, and local context. S2R first maps the member tokens of two candidate regions through an encoder, then exchanges information between the regions through bidirectional attention. A multilayer perceptron (MLP) produces a merge score for the region pair. The model uses member observations and local layout to decide whether candidate regions should merge, and combines regions along accepted connections. Spatial relations accumulated earlier thus inform the membership of functional regions. Direct neighbors supply context to member tokens but do not become members of the region merely because their information is read. Appendix~\ref{app:s2r-details} details token encoding, input processing, and candidate organization.

\subsubsection{Multi-Scale Organization through Spatial and Functional Relations}

Spatial parsing assigns functional meaning to regions and connects local functional spaces into larger spatial structures. After S2R determines region membership, Qwen-VL~\cite{qwen3vl2025} uses the spatial layout and semantic labels of map nodes to generate region names and descriptions. The system proposes candidate combinations based on spatial relations between regions, and Qwen-VL determines whether their functional connections support further grouping. These decisions draw on local top-down views, object information supplied by member nodes, and inter-region relations. Accepted combinations create parent regions with names, descriptions, and links to their subregions. Regions that are not merged retain their members and remain accessible spatial units. The resulting hierarchy connects object composition, local function, and broader spatial meaning: higher-level descriptions summarize spatial context, while subregion and membership links preserve paths to local details. Together, they provide connected evidence at multiple scales for subsequent goal selection.

Higher-level integration treats existing regions as compositional units, reusing their names, descriptions, and subregion composition to determine whether spatial and functional relations support a larger region. If the model rejects all candidates, the existing organization is retained; if it accepts some, corresponding parent nodes are added. The small, medium, and large scales in Figure~\ref{fig:method-overview} therefore denote spatial extents progressively composed from local regions, rather than predefined architectural categories. Figure~\ref{fig:method-architecture}(c) shows how hierarchical links connect these extents. A functional region may lie within a single room, while larger combinations remain grounded in spatial and functional relations. The hierarchy can consequently represent spatial meaning both within rooms and across larger extents.

Each periodic semantic parsing update reprocesses changed regions and replaces their spatial interpretations. When a subregion changes, the system also reinterprets and replaces the function and description of its parent region, while retaining the results for unchanged subregions. These replacements affect only the semantic parsing output; they do not alter online map nodes or the first DF Louvain partition. The most recent available hierarchy remains in use until the next parsing result is ready. This update mechanism reuses interpretations of unchanged regions to reduce repeated parsing while allowing the online map to continue incorporating observations. Region and membership links preserve access from higher-level functional descriptions to underlying map nodes, providing the basis for selecting spatial and object goals from language instructions.

\subsection{PFC-Inspired Multi-Scale Goal-Directed Navigation}
\label{sec:navigation}
\subsubsection{Instruction-Driven Evidence Selection across Scales}

The region hierarchy connects object details to spatial context, enabling the navigation module to select goals according to the instruction. Inspired by the role of goals and context in hierarchical behavioral control in the prefrontal cortex (PFC)~\cite{koechlin2003control}, the system combines region semantics, local composition, and object evidence to make decisions. The upper part of Figure~\ref{fig:method-architecture}(a) connects spatial understanding to goal selection: the instruction determines which evidence is needed, and hierarchical and membership links ground the selection in a map location.

Given a language instruction, Qwen3-VL~\cite{qwen3vl2025} uses the semantic hierarchy to select candidate spaces and gathers additional evidence through region, node, and relation queries. For example, when searching for a cup in a particular functional region, region semantics supply the spatial context, while cup attributes and relations to reference objects distinguish candidates. Queries combine information at different scales as needed by the instruction. They need neither follow a fixed scene--room--region--object order nor include every detail from every level in the decision. The system ultimately identifies a target space linked to the map; if the evidence is insufficient, it does not return a target.

\subsubsection{On-Demand Relation Evidence from 3D Gaussian Representations}

When region semantics and object attributes do not adequately distinguish candidates, the system queries spatial relations between nodes as needed, for example to check whether a cup is on a table. Each node is represented by multiple three-dimensional Gaussian kernels fitted to its observed geometry. A separate geometric relation model takes the representations of a pair of nodes and predicts four relations: \texttt{on}, \texttt{in}, \texttt{overlap}, and \texttt{near}. These predictions supply additional local geometric evidence for comparing targets. They serve as reference evidence and do not automatically eliminate candidates. Appendix~\ref{app:navigation-relations} describes the model inputs and prediction procedure.

On-demand relation queries compute evidence for node pairs relevant to the current goal, avoiding the need to enumerate all object relations in advance. They use three-dimensional Gaussian representations directly to support goal selection. Online GSIF connections, in contrast, accumulate sparse relations from observations to support region partitioning and S2R neighborhood context. The two mechanisms serve goal selection and map organization, respectively.

\subsubsection{Goal Grounding and Navigation Execution}

The system converts the selected semantic target into a navigation endpoint in nearby free space. For an object goal, it uses the selected instance's map geometry to choose the nearest free-space point in its vicinity. For a region goal, it selects the largest free-space component within the region that is connected to the exterior and determines an endpoint within that component.  Appendix~\ref{app:navigation-endpoints} describes endpoint selection and the handling of unreachable targets.

\subsection[Implementation Details]{Implementation Details}
\label{sec:implementation-details}

\textbf{Language model quantization.} Spatial parsing and navigation use Qwen3-VL-8B-Thinking. Its language module uses GGUF Q4\_K\_M quantization and occupies $4,455.34\,\mathrm{MiB}$. The visual encoder and projection module use Q8\_0 quantization and occupy $717.42\,\mathrm{MiB}$. The model weights occupy approximately $5.05\,\mathrm{GiB}$ of GPU memory.

\textbf{S2R node encoding.} A node's own input concatenates $256$-dimensional observation features, its three-dimensional center, and its three-dimensional scale, yielding $262$ dimensions. The normalized center and scale describe the node's position and extent. The model first encodes this input, then aggregates one layer of neighbor information along the sparse connections accumulated by GSIF. Neighbor messages combine connected-node encodings, three-dimensional relative positions, and the most recently recorded GSIF cost for each connection to produce a $128$-dimensional context. A gate controls the context contribution before it is concatenated with the node's own input, yielding a $390$-dimensional node token. Nodes without neighbors use a zero context and retain their own observations and geometry. Statistics for center normalization and cost standardization are determined from the training set alone and remain fixed during inference; scales and relative displacements use a common fixed unit of length. A region remains a set of member-node tokens, and neighbor aggregation does not change membership. Appendix~\ref{app:s2r-details} provides the encoding equations. S2R weights remain fixed when the model is transferred to HM3D.

\textbf{Inputs across rounds of spatial hierarchy parsing.} The first two rounds combine local top-down views, object information from member nodes, and inter-region relations to determine whether functional connections support grouping the candidate regions. From the third round onward, the model uses existing region names, descriptions, and subregion composition to assess further merging. Accepted combinations form parent nodes with names, descriptions, and links to subregions; unmerged regions remain accessible units. Parsing proceeds to the next round if any combination is accepted and terminates when all candidates are rejected. Combinations that have been explicitly rejected are not resubmitted in later rounds.

\section{Experiments}
\label{sec:experiments}

\subsection{Experimental Setup}
\label{sec:exp-protocol}
We evaluate EMS-Nav on 36 scenes from the HM3D validation split~\cite{hm3d2021} for language-guided navigation across spatial scales. The system builds a scene representation along existing camera trajectories, then selects a destination from a language instruction and navigates to it. Mapping takes registered RGB-D observations, camera intrinsics, and known poses as input. All methods receive the same 47 observation trajectories from the HM3D validation split, obtained from the public \href{https://github.com/RuiZhou-cn/memory-over-maps}{Memory Over Maps repository}~\cite{memoryovermaps}. Sampling these trajectories at a fixed stride yields 26,647 input frames. Both the main comparison and the ablation study use the full LangMap instruction set~\cite{miao2026langmap}: 18,628 instructions spanning object, room, region, and instance constraints. We compare against ConceptGraphs (CG)~\cite{conceptgraphs}, HOV-SG, and DualMap~\cite{hovsg,dualmap}. RGB-D rendering and subsampling settings are provided in Appendix~\ref{app:exp-details}.

We measure cumulative navigation success rate (SR). An instruction is successful if the agent reaches a location within $1\,\mathrm{m}$ of the target at least once in a budget of $k$ navigation attempts. All methods use the same attempt strategy, and we report SR@1, SR@3, and SR@5, corresponding to budgets of one, three, and five navigation attempts, respectively. Navigation is executed in Habitat-Sim~\cite{habitat2019} using the selected free-space endpoint. The stopping radius is separate from the success threshold; endpoint selection, motion settings, and unreachable-target handling are detailed in Appendix~\ref{app:navigation-endpoints}. The main experiments run on an NVIDIA RTX 4090. EMS-Nav uses FastSAM and ESAM~\cite{fastsam2023,esam2025} in its mapping front end and MobileCLIP-S2~\cite{mobileclip2024} for semantic features. Its default language reasoning model is a quantized Qwen3-VL-8B-Thinking~\cite{qwen3vl2025}. HOV-SG uses the same language model as EMS-Nav, CG uses Qwen3.7-plus, and DualMap does not require a language model. The real-world deployment uses an RTX 4060 Ti; its hardware and task settings are described in the deployment experiment.

Within each trajectory, observations are sampled at a fixed stride of 5, with the corresponding RGB images, depth maps, and poses read together. Images are saved at $640\times480$ resolution. The instruction set includes both concise and detailed language styles, and the main results aggregate both styles.

\subsection{Quantitative Results: Navigation and Computational Cost}
\label{sec:exp-navigation}
\subsubsection{Navigation across Spatial Scales}
\begin{figure}[tbp]
\centering
\includegraphics[width=0.8\linewidth]{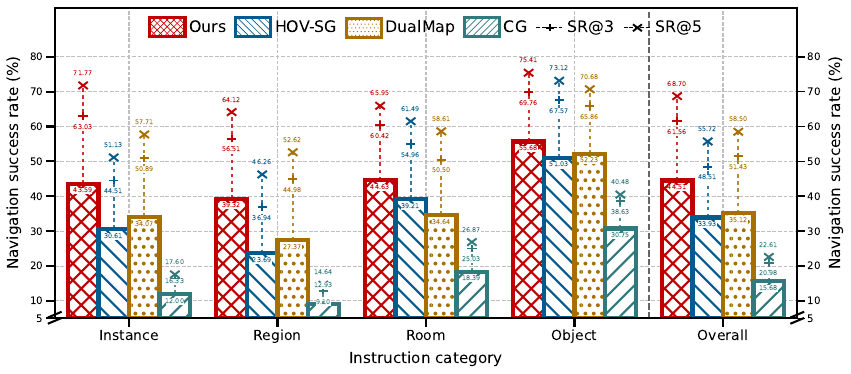}
\caption{Cumulative navigation success under object, room, region, and instance constraints, together with overall results. Bars show SR@1; plus and cross markers indicate SR@3 and SR@5, respectively. The vertical axis contains a break. EMS-Nav leads across all four goal categories and attempt budgets, showing that one environment representation supports language goals at different spatial scales.}
\Description{Cumulative navigation success rates for EMS-Nav, HOV-SG, DualMap, and CG, shown for four instruction categories and overall.}
\label{fig:exp-navigation}
\end{figure}

EMS-Nav achieves the highest cumulative navigation success rate for all four types of goal constraints (Figure~\ref{fig:exp-navigation}). Its overall SR@1, SR@3, and SR@5 reach 44.51\%, 61.56\%, and 68.70\%, respectively, exceeding DualMap, the strongest baseline overall, by 9.39, 10.13, and 10.20 percentage points. EMS-Nav maintains this lead across attempt budgets, from a single attempt to as many as five. Supporting all four types of constraints with one environment representation demonstrates its ability to connect objects, functional spaces, and the broader environment for multi-scale navigation.

Increasing the attempt budget from one to three raises EMS-Nav’s success rate by 17.05 percentage points, while increasing it from three to five adds another 7.14 points. Additional attempts therefore recover some initially unsuccessful tasks, although the gains diminish as the budget grows.

EMS-Nav also retains substantial advantages under region and instance constraints. Its region and instance SR@1 scores are 39.32\% and 43.59\%, exceeding the strongest respective baselines by 11.95 and 9.52 percentage points. At SR@5, these scores reach 64.12\% and 71.77\%, with corresponding gains of 11.50 and 14.06 percentage points. Compared with goals that specify only an object category, region and instance constraints require the system to distinguish a local space or a particular object. These results show that EMS-Nav's advantage extends to more specific goal constraints. The ablation study further examines the respective contributions of functional regions and multi-scale organization.

\subsubsection{Mapping Quality and Efficiency as Scene Size Increases}
\label{sec:exp-mapping}
Mapping quality is evaluated through both class-agnostic and semantic instance reconstruction. Each method's final instance map is projected onto the same fixed set of points sampled from RGB-D observations. Predicted and ground-truth instances are matched using point-set intersection over union (IoU). Average precision (AP) is averaged over nine IoU thresholds from $0.50$ to $0.90$ in increments of $0.05$; AP50 and AP25 use thresholds of $0.50$ and $0.25$, respectively. A match requires an IoU strictly greater than the threshold. Class-agnostic metrics are averaged arithmetically across the 36 scenes. Semantic evaluation uses the MPCat40 taxonomy and additionally requires matching predicted and ground-truth classes. Results are pooled across scenes within each class, then macro-averaged over classes with valid ground truth. The full point-mapping and class-filtering rules are provided in Appendix~\ref{app:exp-details}.

\begin{figure}[t]\centering
\includegraphics[width=\linewidth]{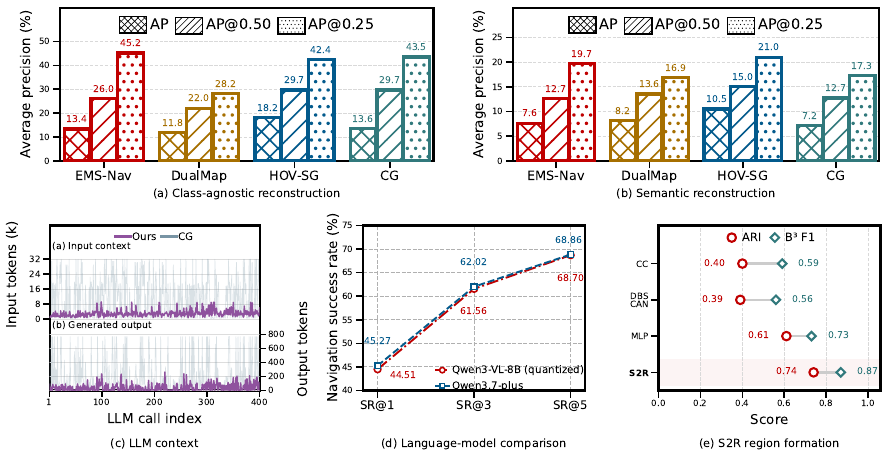}
\caption{Perceptual quality and component comparisons for multi-scale navigation. (a,b) Class-agnostic and semantic instance reconstruction. (c) Language-model input/output tokens across online mapping and querying. (d) Navigation with quantized Qwen3-VL-8B-Thinking and Qwen3.7-plus. (e) Region grouping quality measured by ARI and B$^3$ F1. Token curves follow model-call order; one isolated EMS-Nav output spike is median-smoothed for display.}
\Description{Top row, left to right: class-agnostic instance reconstruction accuracy, and semantic instance reconstruction accuracy. Bottom row: language-model input and output tokens, navigation success rates with quantized Qwen3-VL-8B-Thinking and Qwen3.7-plus, and ARI and B-cubed F1 for S2R and alternative region grouping methods. Token curves follow model-call order, with one isolated EMS-Nav output spike median-smoothed for display. CC is distance-based connected components, DBSCAN is density-based clustering, MLP is a multilayer perceptron, and S2R is our region formation method.}

\label{fig:exp-combined}
\label{fig:exp-mapping}
\label{fig:exp-efficiency}
\label{fig:exp-generalization}
\label{fig:exp-s2r}
\end{figure}

The lightweight mapping front end preserves instance reconstruction quality close to CG while achieving higher semantic AP and class-agnostic AP25 (Figure~\ref{fig:exp-combined}(a,b)). EMS-Nav achieves a class-agnostic AP of 13.36, close to CG's 13.63, and a semantic AP of 7.63, above CG's 7.15. Its class-agnostic AP25 reaches 45.23, exceeding CG's 43.53 and DualMap's 28.15. It also improves on DualMap in class-agnostic AP and AP50, from 11.84 and 22.03 to 13.36 and 26.01, respectively. Although EMS-Nav's class-agnostic and semantic AP remain below HOV-SG's 18.23 and 10.53, it achieves higher navigation success rates for all four goal types. Instance reconstruction accuracy alone therefore does not capture how well a map supports multi-scale navigation. We further examine the roles of functional regions and spatial hierarchies in the ablation study.

In the tests shown, EMS-Nav's per-frame mapping time grows slowly with input sequence length (Figure~\ref{fig:exp-construction}(b)). The timing covers computation through fine-grained region generation, excluding language-model parsing for semantic hierarchy construction. For original sequences of 2k, 3k, and 4k frames, the fitted curves estimate processing times of 176, 189, and 199\,ms per frame, respectively. Across these fitted estimates, from 2k to 4k, this cost increases by approximately 13.1\%, whereas DualMap increases from 421 to 713\,ms per frame and CG from 877 to 1,090\,ms per frame. At 4k, EMS-Nav reduces processing time by approximately 72.1\% and 81.7\% relative to these two methods.  This behavior is consistent with using regions to bound local updates and reuse existing spatial organization. Together with the navigation results, these measurements show higher success across spatial scales and lower per-frame mapping times under the tested settings.

On an RTX 4060 Ti, measurements across the mapping runs and all 18,628 navigation instructions give a mean hierarchical parsing time of 4.32\,s per update and a mean navigation-query time of 3.97\,s.

\subsubsection{Lightweight Perception and Compact Semantic Input}
\label{sec:exp-efficiency}
\begin{figure}[t]\centering
\includegraphics[width=\linewidth]{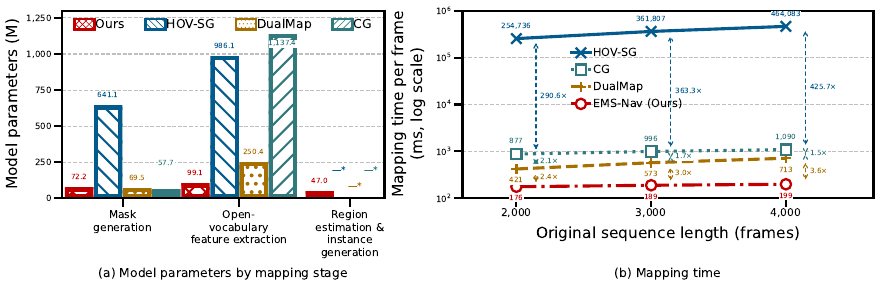}
\caption{Mapping model size and processing cost. (a) Parameter counts for three mapping stages; asterisks indicate no additional learned module. (b) Fitted per-frame mapping-time estimates at original sequence lengths of 2k, 3k, and 4k frames. Dashed double-headed arrows show processing-time ratios between adjacent methods; the vertical axis is logarithmic. EMS-Nav maintains low mapping costs with modest growth over the tested range.}
\Description{Left: model parameter counts for three mapping stages. Right: per-frame mapping times and relative processing-time ratios for four methods at three input sizes.}
\label{fig:exp-construction}
\end{figure}

EMS-Nav reduces the parameter count of its mapping front end through smaller perception models (Figure~\ref{fig:exp-construction}(a)). Mask acquisition uses 72.2M parameters, open-vocabulary feature extraction uses 99.1M, and region estimation and instance generation use 47.0M, totaling approximately 218.3M across the three mapping stages. HOV-SG uses 641.1M and 986.1M parameters for mask and feature acquisition, respectively, while CG's feature acquisition stage uses 1,137.4M. Together with the reconstruction results in Figure~\ref{fig:exp-combined}(a,b) and the navigation results, these comparisons show that smaller perception models can still provide the instance information needed for spatial parsing, combining multi-scale navigation with a lightweight front end.

EMS-Nav also keeps language-model inputs compact throughout online mapping and querying (Figure~\ref{fig:exp-combined}(c)), including both spatial parsing and navigation-query calls. Its inputs are mostly on the order of a few thousand tokens, whereas CG's repeatedly approach 32,000 tokens. Region-level semantic organization summarizes local object compositions as functional names and descriptions. Subsequent spatial composition reuses these summaries without repeatedly expanding all underlying object details. This result is consistent with the design of region summarization and reuse: regions encode the spatial meaning needed for navigation while supplying compact inputs for understanding larger spaces. They thus connect semantic organization with the organization of computation.

\subsection{Qualitative Analysis: Functional Regions across Spatial Scales}
\label{sec:exp-qualitative}
\begin{figure}[tbp]\centering
\includegraphics[width=\linewidth,pagebox=cropbox]{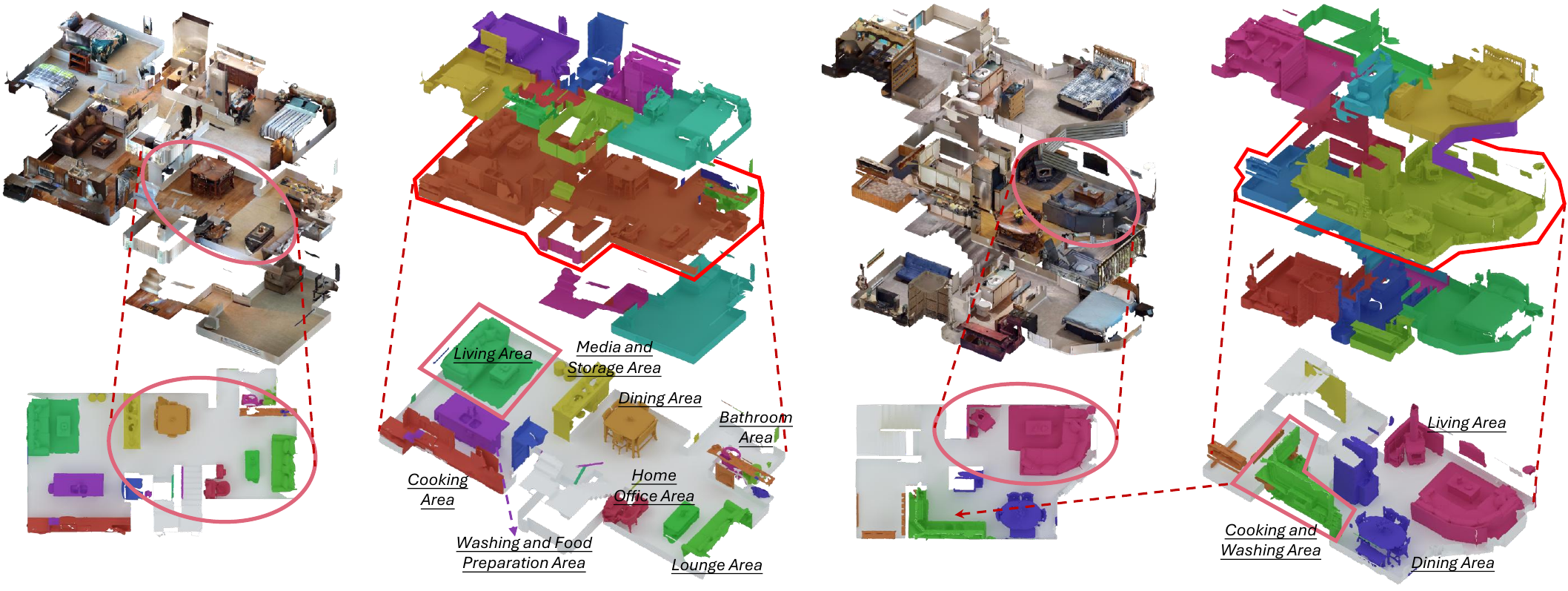}
\caption{Multi-scale functional regions in HM3D scenes \texttt{qyAac8rV8Zk} (left) and \texttt{mL8ThkuaVTM} (right). Each example shows the original scene, its large regions, and functional regions within an open space; lines link corresponding local views. Living, dining, cooking, and other functions remain distinguishable within the larger space, illustrating spatial organization beyond room boundaries.}
\Description{Two scenes show functional regions within local open spaces, alongside the original three-dimensional observations and floor-level regions. Red lines link corresponding local views.}
\label{fig:exp-qualitative}
\end{figure}

Functional differences within open-plan spaces require a representation finer than a room or a broad region. Both examples in Figure~\ref{fig:exp-qualitative} first show the scene and its floor-level partition into large regions, followed by local views of functional regions within an open space. The left example distinguishes living, dining, cooking, washing, and food-preparation areas within a larger open space, while also retaining local functions such as home-office work, leisure, and media and storage. The correspondence between the larger regions and these functional areas illustrates how the same space is organized at different scales.

In the left scene, \texttt{qyAac8rV8Zk}, the functional regions capture differences in use between adjacent activity spaces. Cooking and washing/food preparation occupy the same open kitchen, yet are organized separately, preserving finer functional distinctions within it. This allows the representation to express both the larger region and its local activity spaces: the former maintains the continuity of the open space, while the latter distinguishes where cooking and washing/food preparation take place. Even without separate room boundaries, the map therefore provides spatial references tied to specific uses, grounding functional descriptions in local parts of the open space.

The right scene, \texttt{mL8ThkuaVTM}, shows similarly clear functional organization in a different open-plan layout. First, the space enclosed by sofas and the nearby dining table and chairs form separate living and dining regions. Although both belong to the same larger region, the local partition preserves the distinction between relaxing and dining, allowing ``the dining area in the open space'' to refer to a more specific extent. Second, cooking and washing functions along the kitchen workspace are organized into one region, preserving this continuous activity space as a whole. In contrast to their separate representation in the left scene, this example illustrates how functional organization captures different local combinations across scenes. Together, these examples show that EMS-Nav forms meaningful functional regions even in open spaces without clear room divisions. Upper-level regions preserve the overall spatial extent, while lower-level regions distinguish local uses, providing context for goals that refer to functional spaces.

\subsection{Transfer across Scenes and Language-Model Compatibility}
\label{sec:exp-generalization}
\subsubsection{Transfer without Target-Scene Fine-Tuning}
EMS-Nav transfers to HM3D scenes with fixed models and strategies, and handles navigation instructions supplied after mapping. ESAM and S2R are trained on ScanNet, while FastSAM, MobileCLIP, and the vision-language model are used as general-purpose pretrained components. No additional training is performed on HM3D scenes or evaluation instructions. The main results across all four constraint types show that the system can handle these navigation goals in the evaluated HM3D scenes without target-scene fine-tuning.

\subsubsection{Performance across Language Models}
A quantized 8B language model already supports high navigation success, with limited gains from replacing it with Qwen3.7-plus (Figure~\ref{fig:exp-combined}(d)). The two configurations achieve overall SR@1 scores of 44.51\% and 45.27\%, SR@3 scores of 61.56\% and 62.02\%, and SR@5 scores of 68.70\% and 68.86\%, respectively. The differences are 0.76, 0.46, and 0.16 percentage points. On this evaluation set, the quantized 8B model approaches the navigation performance of the replacement model, supporting its use in the subsequent real-world deployment.

The quantization configuration of the local model is provided in Section~\ref{sec:implementation-details}.

The map supplies either language model with region semantics, object evidence, and links between spatial scales for interpreting an instruction. The similar success rates are consistent with this structured context supporting target selection with the quantized model.

\subsection{Real-World Deployment}
\label{sec:exp-deployment}
We deploy EMS-Nav in real indoor environments spanning offices, teaching spaces, laboratories, and shared activity spaces. Observations are collected with an Orbbec Femto Bolt RGB-D camera, and computation runs on an NVIDIA RTX 4060 Ti. The camera configuration and other hardware and software details are provided in Appendix~\ref{app:exp-details}. Figure~\ref{fig:exp-deployment}a shows a selection of the environments and collection routes.

The deployment evaluation comprises 128 navigation instructions, with all targets observed by the camera before the corresponding query is issued. Following DualMap~\cite{dualmap}, we use Habitat Data Collector for scene-data collection and the same ROS2--Habitat navigation configuration. Each instruction permits one attempt. ROS2 connects the navigation system to Habitat, and Habitat executes the actions in the captured scene. The reached space is then checked against the instruction. The attempt is successful if the space satisfies the instruction and unsuccessful otherwise. Under this protocol, the single-attempt success rate is 61.7\%. The evaluation covers bookshelves, drinking-water areas, corridors, desks, offices, classrooms, laboratories, meeting rooms, common spaces, and restrooms. These results demonstrate EMS-Nav's ability to ground language instructions in target spaces using real observations across diverse spatial contexts.

The system completes the on-site evaluation on the RTX 4060 Ti without runtime interruptions. GPU memory usage remains at approximately 8\,GiB in Figure~\ref{fig:exp-deployment}b, and peak runtime throughput reaches 5.19 FPS. Hardware and execution details are provided in Appendix~\ref{app:exp-details}. Together, these runtime records and target-space selection results demonstrate EMS-Nav's ability to integrate mapping, spatial understanding, and language-goal processing on a single local platform, making its multi-scale environment representation usable for real indoor tasks.

Over the approximately 400 seconds shown in Figure~\ref{fig:exp-deployment}b, GPU utilization remains mostly high and memory usage shows no sustained upward trend. The trace indicates stable memory demand while the local platform continues processing environmental observations.

\begin{figure}[tbp]
\centering
\begin{minipage}[t]{0.530\linewidth}\vspace{0pt}\centering
\parbox[c][2.0606in][c]{\linewidth}{\centering\includegraphics[width=\linewidth]{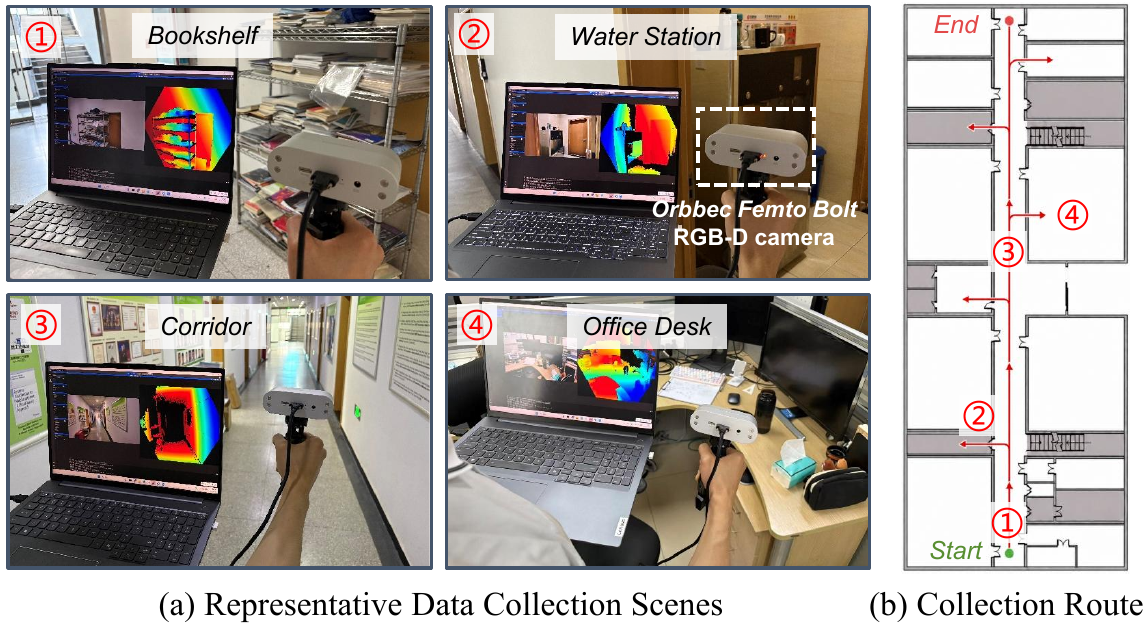}}\par
\vspace{2pt}{\small\centering (a) Environments, equipment, and collection routes\par}
\end{minipage}\hspace{.008\linewidth}\begin{minipage}[t]{0.462\linewidth}\vspace{0pt}\centering
\parbox[c][2.0606in][c]{\linewidth}{\centering\includegraphics[width=\linewidth]{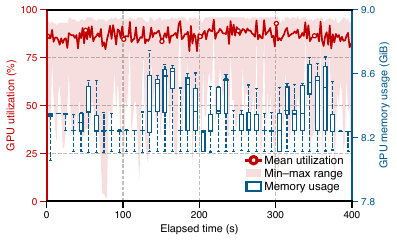}}\par
\vspace{2pt}{\small\centering (b) GPU utilization and memory usage\par}
\end{minipage}
\caption{Real-world deployment and resource measurements on an RTX 4060 Ti. (a) Representative indoor environments, the RGB-D device, and collection routes illustrate the observation settings. (b) GPU utilization and memory usage during operation show the computational workload and memory demand of local processing. These measurements complement the navigation results reported in the text when assessing on-device operation.}
\Description{Deployment environments are shown alongside GPU utilization and memory-usage traces.}
\label{fig:exp-deployment}
\end{figure}

\subsection{Ablation Study}
\label{sec:exp-ablation}

\begin{figure}[tbp]
\centering
\includegraphics[width=0.8\linewidth]{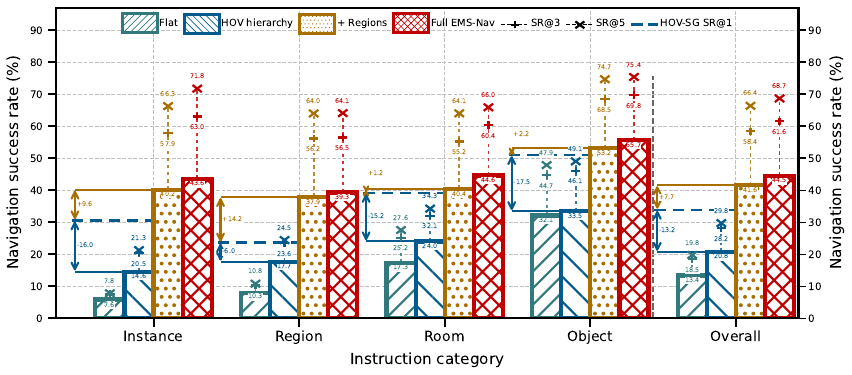}
\caption{Ablation of functional regions and multi-scale organization with a shared instance map. Bars show SR@1 on the left axis; plus and cross markers show SR@3 and SR@5 on the right axis. Blue dashed lines mark original HOV-SG SR@1; blue and ochre arrows show differences from this reference. The four settings separate the contributions of functional regions and higher-level organization. Labels are rounded to one decimal place.}
\Description{A closed-frame plot with two vertical axes contains all SR@1 bars, SR@3 plus signs, SR@5 cross markers, blue HOV-SG reference lines, and ten groups of difference arrows.}
\label{fig:exp-ablation}
\end{figure}

\subsubsection{Contributions of Functional Regions and Multi-Scale Organization}
Here, HOV-SG-style organization uses the room and floor information from HOV-SG to organize the shared instance map. We compare four spatial organizations built on EMS-Nav's instance representation: a flat instance map; an instance map with HOV-SG-style multi-scale organization; the same organization augmented with EMS-Nav functional regions; and an instance map with functional regions organized using EMS-Nav's multi-scale hierarchy. Their overall SR@1 scores are 13.42\%, 20.78\%, 41.63\%, and 44.51\%, respectively, and their SR@5 scores are 19.83\%, 29.76\%, 66.40\%, and 68.70\% (Figure~\ref{fig:exp-ablation}). Moving from flat instances to HOV-SG-style organization improves SR@1 by 7.36 percentage points.  Adding functional regions yields a further 20.85-point improvement, identifying region information as an important source of the navigation gains.

With EMS-Nav's instance representation, functional regions recover and then surpass the navigation performance of the original HOV-SG. In the main experiment, the original HOV-SG achieves an overall SR@1 of 33.93\%. Using EMS-Nav instances while retaining HOV-SG-style organization yields 20.78\%. Adding EMS-Nav functional regions raises SR@1 to 41.63\%, exceeding the original HOV-SG by 7.70 percentage points. SR@5 likewise rises from 29.76\% to 66.40\%, above the original HOV-SG's 55.72\%. Together with the mapping results, this comparison shows that organizing objects into functionally meaningful spaces can yield higher navigation success despite lower instance reconstruction accuracy. Functional regions connect instance perception to language goals, enabling a lighter instance map to support more effective spatial understanding and destination selection.

Because the instance map remains fixed when functional regions are added, this recovery cannot be attributed to finer instance reconstruction. The comparison instead supports the contribution of local functional context to destination selection.

Adding functional regions to HOV-SG-style organization improves navigation success for all four goal types, with the largest gain for instance constraints. SR@1 increases by 19.71, 16.37, 20.18, and 25.62 percentage points for object, room, region, and instance goals, respectively. Instance constraints require selecting a particular instance from candidate objects. Their larger gain is consistent with functional regions supplying additional local context.

With the same instance and functional-region representations, EMS-Nav's multi-scale organization further improves navigation success. Relative to HOV-SG-style organization, overall SR@1 increases from 41.63\% to 44.51\%, SR@3 from 58.42\% to 61.56\%, and SR@5 from 66.40\% to 68.70\%. Room SR@1 increases from 40.41\% to 44.63\%. This comparison shows that the spatial organization above functional regions also affects navigation: linking local functional spaces into a multi-scale hierarchy improves destination selection beyond the context already provided by individual regions. Functional regions and their hierarchical organization thus work together to support multi-scale navigation.

\subsubsection{S2R Region Merging}
The region grouping experiment uses 1513 ScanNet scenes, split into 960 for training, 241 for validation, and 312 for testing. Comparison methods include distance-based connected components (CC), density-based clustering (DBSCAN), and a multilayer perceptron (MLP). Supervision regions are initially annotated by a large model and then manually checked; the training data comprise 980,000 sample pairs. All region-merging methods share the initial regions, acyclic candidate generation, score ranking, and grouping along accepted connections. Each performs at most 3 rounds of merging and candidate updates. Thresholds are selected separately for each model on the validation set, using the same search budget and selection objective. Evaluation aligns predictions and annotations to the same $1\,\mathrm{cm}$ voxel domain and computes ARI and B$^3$ F1. Detailed feature settings and voxel-label assignment rules are provided in Appendix~\ref{app:exp-details}.

S2R recovers region membership more accurately than simple distance-based grouping and an MLP alternative (Figure~\ref{fig:exp-combined}(e)). Distance-based connected components and DBSCAN~\cite{dbscan1996} achieve adjusted Rand index (ARI)~\cite{ari1985} scores of 0.40 and 0.39, and B$^3$ F1~\cite{bcubed1998} scores of 0.59 and 0.56, respectively. The MLP baseline reaches 0.61 and 0.73, while S2R further improves these scores to 0.74 and 0.87, gains of 0.13 and 0.14. Both metrics indicate that S2R more closely recovers membership in the supervised regions than the compared alternatives, supporting the formation of functionally meaningful spatial units from the object map.

\section[Limitations and Future Work]{Limitations and Future Work}
\label{sec:limitations}

\textbf{Observation continuity.} EMS-Nav is not suited to observation sequences with excessively large intervals between frames. Instance association and the accumulation of local spatial relations rely on connections established across successive observations. Large observation gaps weaken this evidence, limiting the system's ability to maintain a coherent map and organize local spaces. Extending the framework to low-frequency or discontinuous observations is therefore an important direction for future work, particularly when sensing or communication constraints prevent frequent observations.

\textbf{On-device computational cost.} Although EMS-Nav runs locally, continuous mapping and periodic spatial parsing still incur computational costs that constrain deployment on more resource-limited devices. Further improvements should reduce the cost of maintaining and interpreting the map while preserving the spatial information needed for navigation. Promising directions include lighter perception and region-reasoning models, together with update schedules that adapt to changes in the environment. These improvements would broaden the range of devices on which multi-scale spatial understanding can operate continuously.

\textbf{Semantic reliability and feature dependence.} EMS-Nav forms regions by aggregating map nodes, each of which is a superpoint or an aggregate of superpoints, rather than requiring fully reconstructed object instances. Precise instance boundaries are therefore not a prerequisite for region formation, and boundary accuracy alone does not determine whether the map can identify the space in which a target is likely to be found. However, errors in object semantics can propagate to the interpretation of regional functions and impair spatial retrieval. The current S2R model also depends on features extracted by ESAM, limiting the flexibility to replace the perception front end. Future work will investigate region reasoning that is more robust to semantic errors and compatible with different visual encoders, including general-purpose vision-language features, to reduce dependence on a particular feature representation.

\section{Conclusion}
\label{sec:conclusion}
EMS-Nav unifies multi-scale spatial understanding and language-guided action within a region-centered framework for on-device embodied navigation. Its central contribution is to make regions the organizing basis for perceptual mapping, spatial understanding, and goal-directed decisions, connecting objects, functional spaces, and the broader environment so that robots can interpret and respond to navigation instructions at different scales. This design supports efficient local mapping alongside higher-level spatial reasoning, bringing rich environmental understanding to navigation under limited computational resources. Leading results across four target categories, together with local deployment experiments using real-world observations, demonstrate the value of advancing navigation capability and computational efficiency together. EMS-Nav offers a unified approach spanning environmental perception, spatial understanding, and language-guided action, enabling robots with limited computational resources to turn people's descriptions of rooms, functional regions, and objects into navigation actions.

\bibliographystyle{ACM-Reference-Format}
\bibliography{bib/references}
\clearpage
\appendix
\section{GSIF Candidate Connections and Relation Weights}
\label{app:gsif-relations}

Candidate connections are proposed between nearby nodes observed together. Let $p_i\in\mathbb{R}^{2}$ denote the horizontal position of node $i$ and $q$ a query position in its local field. During a query, the source object's own potential provides local repulsion, directing the field away from the object. Together with attraction from other objects and repulsion from fixed architectural structures, this term defines the query potential $U^{(i)}$.

On the horizontal plane, let $K_j(x)$ denote the nonnegative attraction kernel of object $j$, $W_b(x)$ a structural repulsion kernel, and $P_i(x)$ the source object's own repulsion potential. The source-conditioned potential is
\begin{equation}
U^{(i)}(x)=\sum_b W_b(x)-\sum_{j\ne i}K_j(x)+P_i(x).
\end{equation}
The source object's attraction is excluded, while $P_i$ provides its local repulsion.

The resulting guidance direction is
\begin{equation}
  g^{(i)}(q)=-\nabla U^{(i)}(q).
\end{equation}
The alignment between the field direction and the direction from the query position to candidate node $j$ is
\begin{equation}
  a_{ij}(q)=\frac{g^{(i)}(q)^\top(p_j-q)}{\|g^{(i)}(q)\|\,\|p_j-q\|}.
\end{equation}
A GSIF query assigns the following candidate cost:
\begin{equation}
  s_{ij}(q)=1-a_{ij}(q)+\max\!\left\{0,\,U^{(i)}(p_j)-U^{(i)}(q)\right\}.
\end{equation}
The directional term $1-a_{ij}(q)$ penalizes disagreement with the field direction, while the second term penalizes an increase in potential from the query position to the candidate position. A lower cost indicates a connection that better follows the guidance of that query.
Directional agreement is set to zero when the field direction is degenerate or the candidate node coincides with the query position. The same undirected node pair may be nominated more than once in a frame; the system retains the lowest candidate cost. It then processes candidates in cost order, prioritizing connection coverage subject to a node-degree cap. This procedure does not guarantee that every node reaches the target degree.

Let $E_t$ be the set of undirected edges selected in frame $t$. While node identities remain unchanged, accumulated relations and their support counts are updated as
\begin{equation}
  \mathcal{E}_t=\mathcal{E}_{t-1}\cup E_t,\qquad
  n_{ij,t}=n_{ij,t-1}+\mathbf{1}[\{i,j\}\in E_t].
\end{equation}
Here, $n_{ij,t}$ counts the frames up to $t$ in which the node pair has been selected as a connection.
A node pair receives at most one additional unit of support per frame, and only connections selected into $E_t$ contribute; co-occurrence alone does not increase the count. The per-frame degree cap applies only to connections selected in that frame, not to node degrees in the relation graph accumulated across frames.

For a valid node pair in the accumulated relations, let $\hat f_i$ denote the unit-normalized visual feature of node $i$ and define
\begin{equation}
  v_{ij}=\operatorname{clip}\!\left(\frac{\hat f_i^\top\hat f_j+1}{2},\,0,\,1\right),\qquad
  d_{ij}=\|p_i-p_j\|_2.
\end{equation}
The relation weight is
\begin{equation}
  w_{ij,t}=\left(0.3\,v_{ij}+0.7\,e^{-d_{ij}/(2\,\mathrm{m})}\right)
  \left(0.5+0.5\,\frac{n_{ij,t}}{n_{ij,t}+2}\right).
\end{equation}
Visual similarity and horizontal center distance jointly determine the base weight, which is modulated by a saturating factor for support across frames. The GSIF cost is used to select candidate connections; it is neither substituted directly into this weight nor interpreted as the probability that two nodes belong to the same region.

\section{S2R Representation and Candidate Organization}
\label{app:s2r-details}
\subsection{Node Inputs and Neighbor Encoding}

Section~\ref{sec:implementation-details} specifies the node input configuration; here we give the encoding equations. Let $f_i$, $p_i$, and $b_i$ denote the observed feature, three-dimensional center, and scale of node $i$, respectively. The base input is $x_i=[f_i\Vert\bar p_i\Vert\bar b_i]$, where bars denote normalized geometric quantities, and its encoding is $h_i=f_\theta(x_i)$.

Neighbor messages combine the connected node's encoding, its three-dimensional relative position, and the GSIF candidate connection cost. For $j\in\mathcal N(i)$, let $d_{ij}=[\overline{p_j-p_i}\Vert\bar s_{ij}]\in\mathbb{R}^{4}$, where $s_{ij}$ is the most recently recorded candidate cost for that historical connection. A lower cost indicates better agreement with the field guidance at that query; it is distinct from both a same-region probability and a Louvain edge weight. Node encodings and edge descriptors jointly determine the messages and aggregation weights:
\begin{equation}
  m_{ij}=\phi(h_j,d_{ij}),\qquad
  \alpha_{ij}=\operatorname{softmax}_{j\in\mathcal N(i)}\psi(h_i,h_j,d_{ij}),\qquad
  c_i=\sum_{j\in\mathcal N(i)}\alpha_{ij}m_{ij}.
\end{equation}
A single neighbor-aggregation layer encodes the connected nodes into a $128$-dimensional context vector $c_i$. A gate then adjusts its contribution to the final token:
\begin{equation}
  g_i=\sigma\bigl(\eta(h_i,c_i)\bigr),\qquad
  t_i=[x_i\Vert g_i\odot c_i]\in\mathbb{R}^{390}.
\end{equation}
Existing connections across regions also participate in aggregation, but receiving neighbor context does not change a node's region membership. Region $r$ is still represented by its member tokens, $T_r=\{t_i:i\in r\}$, and message passing itself does not fuse map nodes.

\subsection{GSIF Connection Costs and Input Processing}
The GSIF candidate connection cost combines directional agreement with the increase in potential; a lower value indicates better agreement with the guidance of that query. For each historical edge, the latest recorded cost, implemented as \texttt{last\_score}, is used under the same rule during training and inference. Existing edges must have valid costs: missing values cannot be filled with zero, because a genuine zero cost has a specific meaning. Nodes without neighbors use zero context without constructing additional connections.

Node centers, scales, and relative displacements use consistent map coordinates and length units. Relative displacement is computed from node centers in world coordinates, rather than by subtracting centers expressed in separately recentered regional coordinate systems. Nodes and edge endpoints correspond to the current map state when semantic parsing begins. GSIF proposes connections in the horizontal plane, while three-dimensional centers, scales, and relative positions come from the nodes' observed geometry. The three-dimensional token inputs do not alter the definition of the horizontal GSIF field.

\subsection{Region-Pair Candidates and Grouping}
S2R evaluates candidate region pairs organized without cycles. Candidate connections may share a region: for example, $A$--$B$ and $B$--$C$ can be retained without adding $A$--$C$. S2R scores each retained pair, and accepted connections jointly determine group membership. Accepting both $A$--$B$ and $B$--$C$ therefore groups all three regions together without a separate decision on $A$--$C$. Rejected connections do not trigger a direct merge. This acyclic organization avoids redundant comparisons within an already connected candidate group.

\subsection{Round-Specific Inputs for Spatial Hierarchy Parsing}
\label{app:hierarchy-rounds}

The round-specific inputs and stopping rules for spatial hierarchy parsing are specified in Section~\ref{sec:implementation-details}.

\section{Geometric Relation Prediction for Navigation}
\label{app:navigation-relations}

The geometric relation model takes sets of local three-dimensional Gaussian kernels for a pair of map nodes as input. Each kernel's center and full covariance are estimated from the first- and second-order moments of point coordinates within an adaptive spatial grid cell, and its weight is the fraction of points in that cell. The number of kernels in the full representation varies with the observed geometry; the relation network deterministically selects at most 128 kernels from this representation. The two sets share a reference origin, preserving relative positions, actual scales, and the roles of the source and reference objects.

A shared encoder extracts features from the Gaussian kernels, and a Transformer~\cite{vaswani2017attention} jointly encodes interactions between the two sets. Weighted pooling and max pooling are applied to each node's encodings to form a node-pair representation, from which the model produces scores and decisions for four relation types. Relation prediction is invoked on demand during target selection. Its outputs provide external geometric evidence for the vision-language model to check the instruction, complementing region semantics and object attributes in the target decision.
For an ordered object pair $(A,B)$, \texttt{on} denotes that $A$ is on $B$, and \texttt{in} denotes that $A$ is inside $B$. They are supervised by the standing on/lying on and standing in/lying in/hanging in labels in 3DSSG~\cite{sg3d2020}, respectively. Here, \texttt{overlap} specifically denotes contact or coincidence between observed surfaces, defined as a minimum distance of at most $0.02\,\mathrm{m}$ between the two observed point sets. The \texttt{near} label uses the same distance measure with a threshold of $0.25\,\mathrm{m}$. Each of the four scores is passed through a sigmoid and thresholded independently to produce multi-label predictions. Thresholds maximize per-class F1 on the development set and remain fixed during testing. The outputs are not constrained to be mutually exclusive or symmetric, nor is \texttt{overlap} required to imply \texttt{near}.

The relation model uses geometry from 3RScan and relation supervision from 3DSSG. Training, development, and test sets are split by disjoint physical-scene reference groups, containing 313, 24, and 48 scenes and 86,188, 7,654, and 9,061 ordered object pairs, respectively. The model first undergoes three epochs of weakly supervised pretraining for \texttt{on}/\texttt{in} on 6,436 object pairs from 128 ScanNet scenes, followed by four-label training with per-class weighted binary cross-entropy and label masking. Checkpoints are selected by the minimum per-class F1 on the development set, with macro-averaged F1 used to break ties.

\section{Navigation Endpoints and Unreachable Destinations}
\label{app:navigation-endpoints}

For an object target, the navigation endpoint is the free-space point closest to the predicted instance surface that provides sufficient obstacle clearance. For a region target, the system first selects the largest free space within the region that is connected to the outside, then chooses the point in its safe interior closest to its area centroid. Map coordinates are converted to Habitat coordinates by $(x,y,z)\mapsto(x,z,-y)$. Motion is executed with Habitat-Sim's GreedyGeodesicFollower, using a forward step of $0.05\,\mathrm{m}$, left and right turns of $5^{\circ}$, and no action noise. The agent stops within $0.15\,\mathrm{m}$ of the navigation endpoint, with a total budget of 2000 actions per task. This stopping radius is configured separately from the $1\,\mathrm{m}$ success threshold used in the experiments.

After determining the endpoint, the system attempts to plan a path from the current position to it. If the endpoint is unreachable, the robot moves to the closest reachable position and stops. An attempt fails if the stopping position does not satisfy the evaluation's success criterion. If the attempt budget permits, the model selects the target for the next attempt; the task ends once that budget is exhausted.

\section{Reproducibility Details and Supplementary Results}
\label{app:exp-details}
\subsection{Observation Generation and Fixed Subsampling}
The observation trajectories contain 133,163 poses, divided into 47 segments after an audit of trajectory boundaries. Within each segment, RGB frame identifiers are sorted numerically and subsampled at a fixed index stride of 5 using \texttt{ids[::5]}. Depth images and poses with matching identifiers are retained, yielding 26,647 frames. Habitat-Sim~\cite{habitat2019} renders images at the specified poses at $3200\times2400$, which are then saved at $640\times480$. RGB images are resized using Lanczos interpolation and depth images using nearest-neighbor interpolation. The horizontal field of view is $90^{\circ}$. The saved images have camera intrinsics $f_x=f_y=320$, $c_x=320$, and $c_y=240$. The instruction set includes both concise and detailed language styles, and the main results are aggregated over the complete set.

\subsection{Implementation and Quantization}
The server has two AMD EPYC 7542 processors, approximately $920\,\mathrm{GiB}$ of visible memory, and an NVIDIA RTX 4090 GPU, and runs Ubuntu 22.04.4 LTS and PyTorch 2.3.1+cu121. Section~\ref{sec:exp-protocol} specifies the perception frontends and language models used in the comparisons; Section~\ref{sec:implementation-details} gives the model quantization and cross-dataset inference configuration for S2R.

\subsection{Complete Navigation and Ablation Results}
The main and ablation experiments use the same 47 observation trajectories, the complete set of 18,628 LangMap instructions, and the same navigation attempt strategy. An instruction is successful if the agent reaches within $1\,\mathrm{m}$ of the target location at least once within a budget of $k$ attempts. We report cumulative success rates for $k=1,3,5$. The four ablation settings share the EMS-Nav instance map and respectively use a flat representation, an HOV-SG-style hierarchy, an HOV-SG-style hierarchy augmented with functional regions, and the complete EMS-Nav hierarchy. The original HOV-SG in the main experiment uses its own instance representation and serves only as an external reference for the ablation results.
\begin{table}[!htbp]
\centering
\small
\setlength{\tabcolsep}{2.8pt}
\renewcommand{\arraystretch}{1.02}
\caption{Cumulative navigation success rates (\%) by constraint type and overall.}
\label{tab:nav-full}
\begin{tabular*}{\linewidth}{@{\extracolsep{\fill}}l*{4}{rrr}@{}}
\toprule
 & \multicolumn{3}{c}{EMS-Nav} & \multicolumn{3}{c}{DualMap} & \multicolumn{3}{c}{HOV-SG} & \multicolumn{3}{c}{CG} \\
\cmidrule(lr){2-4} \cmidrule(lr){5-7} \cmidrule(lr){8-10} \cmidrule(lr){11-13}
Constraint & SR@1 & SR@3 & SR@5 & SR@1 & SR@3 & SR@5 & SR@1 & SR@3 & SR@5 & SR@1 & SR@3 & SR@5 \\
\midrule
Object & 55.68 & 69.76 & 75.41 & 52.23 & 65.86 & 70.68 & 51.03 & 67.57 & 73.12 & 30.75 & 38.63 & 40.48 \\
Room & 44.63 & 60.42 & 65.95 & 34.64 & 50.50 & 58.61 & 39.21 & 54.96 & 61.49 & 18.39 & 25.03 & 26.87 \\
Region & 39.32 & 56.51 & 64.12 & 27.37 & 44.98 & 52.62 & 23.69 & 36.94 & 46.26 & 9.10 & 12.93 & 14.64 \\
Instance & 43.59 & 63.03 & 71.77 & 34.07 & 50.89 & 57.71 & 30.61 & 44.51 & 51.13 & 12.00 & 16.33 & 17.60 \\
Overall & 44.51 & 61.56 & 68.70 & 35.12 & 51.43 & 58.50 & 33.93 & 48.51 & 55.72 & 15.68 & 20.98 & 22.61 \\
\bottomrule
\end{tabular*}
\end{table}

\begin{table}[!htbp]
\centering
\small
\setlength{\tabcolsep}{2.8pt}
\renewcommand{\arraystretch}{1.02}
\caption{Complete navigation results (\%) for four spatial organization settings, all using EMS-Nav instance representations.}
\label{tab:ablation-full}
\begin{tabular*}{\linewidth}{@{\extracolsep{\fill}}l*{4}{rrr}@{}}
\toprule
 & \multicolumn{3}{c}{\shortstack{Flat\\instances}} & \multicolumn{3}{c}{\shortstack{HOV-SG-style\\hierarchy}} & \multicolumn{3}{c}{\shortstack{HOV-SG-style\\+ regions}} & \multicolumn{3}{c}{\shortstack{EMS-Nav\\+ regions}} \\
\cmidrule(lr){2-4} \cmidrule(lr){5-7} \cmidrule(lr){8-10} \cmidrule(lr){11-13}
Constraint & SR@1 & SR@3 & SR@5 & SR@1 & SR@3 & SR@5 & SR@1 & SR@3 & SR@5 & SR@1 & SR@3 & SR@5 \\
\midrule
Object & 32.12 & 44.74 & 47.88 & 33.50 & 46.13 & 49.05 & 53.21 & 68.51 & 74.72 & 55.68 & 69.76 & 75.41 \\
Room & 17.31 & 25.20 & 27.61 & 24.04 & 32.06 & 34.29 & 40.41 & 55.24 & 64.06 & 44.63 & 60.42 & 65.95 \\
Region & 8.01 & 10.34 & 10.83 & 17.73 & 23.55 & 24.47 & 37.91 & 56.22 & 64.02 & 39.32 & 56.51 & 64.12 \\
Instance & 5.92 & 7.62 & 7.84 & 14.61 & 20.49 & 21.32 & 40.23 & 57.91 & 66.31 & 43.59 & 63.03 & 71.77 \\
Overall & 13.42 & 18.53 & 19.83 & 20.78 & 28.22 & 29.76 & 41.63 & 58.42 & 66.40 & 44.51 & 61.56 & 68.70 \\
\bottomrule
\end{tabular*}
\end{table}

\subsection{Language Model Substitution and Mapping Quality}

\begin{table}[!htbp]
\centering
\small
\setlength{\tabcolsep}{5pt}
\renewcommand{\arraystretch}{1.02}
\caption{Cumulative navigation success rates (\%) with different language models.}
\label{tab:model-swap}
\begin{tabular*}{\linewidth}{@{\extracolsep{\fill}}lrrr@{}}
\toprule
Language model & SR@1 & SR@3 & SR@5 \\
\midrule
Quantized Qwen3-VL-8B-Thinking & 44.51 & 61.56 & 68.70 \\
Qwen3.7-plus & 45.27 & 62.02 & 68.86 \\
\bottomrule
\end{tabular*}
\end{table}

\begin{table}[!htbp]
\centering
\small
\setlength{\tabcolsep}{4pt}
\renewcommand{\arraystretch}{1.02}
\caption{Complete AP results for instance mapping.}
\label{tab:mapping-full}
\begin{tabular*}{\linewidth}{@{\extracolsep{\fill}}l*{2}{rrr}@{}}
\toprule
 & \multicolumn{3}{c}{Class-agnostic} & \multicolumn{3}{c}{Semantic} \\
\cmidrule(lr){2-4} \cmidrule(lr){5-7}
Method & AP & AP50 & AP25 & AP & AP50 & AP25 \\
\midrule
EMS-Nav & 13.360 & 26.010 & 45.230 & 7.626 & 12.652 & 19.727 \\
CG & 13.630 & 29.700 & 43.530 & 7.150 & 12.740 & 17.340 \\
DualMap & 11.840 & 22.030 & 28.150 & 8.170 & 13.600 & 16.920 \\
HOV-SG & 18.230 & 29.690 & 42.350 & 10.530 & 15.040 & 21.000 \\
EMS-Nav w/o merge & 12.350 & 24.896 & 45.103 & 7.196 & 12.365 & 19.616 \\
\bottomrule
\end{tabular*}
\end{table}

Instance AP measures reconstruction quality within the overlap of predicted and ground-truth coverage. Labels are transferred within $5\,\mathrm{cm}$, and instances with fewer than 100 evaluation points are excluded. Section~\ref{sec:exp-mapping} defines the metrics and aggregation rules.

Semantic evaluation uses the same point support and MPCat40 taxonomy, excluding walls, floors, ceilings, and unlabeled categories. Predicted instances are ranked by their original confidence scores.

\subsection{Region Grouping: Comparison Settings and Evaluation}
The region grouping dataset contains 1513 ScanNet scenes, divided into 960 training, 241 validation, and 312 test scenes, with 980,000 training sample pairs. S2R is trained with binary cross-entropy on region-pair merge decisions. The MLP baseline takes the 256-dimensional ESAM instance features from the two candidate regions and valid candidate edges. Edge attributes comprise a $\log(1+d)$ transformation of reachable distance and a reachability observation indicator; the model outputs a single merge logit. CC and DBSCAN group by distance. S2R instead uses the member tokens defined in the main text and Appendix~\ref{app:s2r-details}, combining each node's own features and geometry with GSIF neighbor context.

To isolate region grouping from map coverage, we define the evaluation domain before applying any grouping method. The shared input map and ground-truth annotations determine a fixed set of jointly covered $1\,\mathrm{cm}$ voxels, with voxel-to-node assignments held constant across methods. Every method assigns all input nodes to regions; unmerged initial regions remain separate groups. Each evaluation voxel inherits its predicted region label from its assigned map node. ARI and B$^3$ F1 are therefore computed over identical voxels without method-dependent filtering. These metrics assess grouping quality on the mapped, annotated support; ground-truth voxels outside the shared map coverage are excluded, so the scores do not measure scene coverage. All comparison methods share the initial regions, acyclic candidate generation, score sorting, grouping along accepted connections, and candidate updates, with at most three merge rounds. Each model's threshold is selected on the validation set using the same threshold-search budget and selection objective.

\textbf{Mapping-time statistic.} Mapping time covers computation through fine-grained region generation and excludes language-model parsing for semantic hierarchy construction. The EMS-Nav curve is fitted to per-scene means over 36 scenes; comparison methods use sequence-length-bin summaries. Values at 2k, 3k, and 4k original frames are fitted estimates. All methods receive the same observation trajectories and fixed subsampling, controlling the input workload. The reported ratios compare fitted estimates at matched original sequence lengths.

\textbf{Parsing schedule.} Spatial parsing is triggered at a configurable frame interval and runs in parallel with online mapping. It processes the current spatial representation while mapping continues to incorporate observations; navigation uses the most recent available semantic hierarchy.

Figure~\ref{fig:exp-combined}(c) plots input and output token counts for language-model calls throughout online mapping and querying, including spatial parsing and navigation queries, in call order.

\subsection{Real-World Evaluation}
Navigation evaluation on the captured real-world scenes follows the ROS2--Habitat setup of DualMap~\cite{dualmap}. We use \href{https://github.com/Eku127/habitat-data-collector}{Habitat Data Collector} for scene-data collection and navigation execution, with the same navigation configuration as DualMap. ROS2 provides communication between the navigation system and Habitat. Habitat executes navigation in the captured scene. Each instruction permits one attempt. The attempt is successful if the reached space satisfies the instruction and unsuccessful otherwise, with no additional attempts. These results are reported separately from the HM3D benchmark, whose success criterion uses a distance threshold.
The Orbbec Femto Bolt RGB-D camera operates at $1280\times960$ and 15 FPS, with its built-in IMU assisting pose estimation. The compute platform comprises an Intel Core i7-14700F processor, 32 GB RAM, a 1 TB PCIe 4.0 SSD, and an RTX 4060 Ti GPU, running Ubuntu 22.04, PyTorch 2.3.1, and CUDA 12.1. We construct 128 navigation instructions for the captured scenes, with all targets observed by the camera before the corresponding query. The manually evaluated navigation success rate is 61.7\%. The deployment covers bookshelves, drinking-water areas, corridors, desks, offices, classrooms, laboratories, meeting rooms, common areas, and restrooms. Peak runtime throughput reaches 5.19 FPS.

\end{document}